\documentclass[a4paper]{article} 

\usepackage[english]{babel} 
\usepackage[T1]{fontenc} 
\usepackage[utf8]{inputenc} 
\usepackage[babel]{microtype} 

\usepackage[margin={0.85in, 0.85in}]{geometry}

\usepackage{cite}
\usepackage{amsmath,amssymb,amsfonts}
\usepackage{algorithmic}
\usepackage{graphicx}
\usepackage{textcomp}
\usepackage{xcolor}
\def\BibTeX{{\rm B\kern-.05em{\sc i\kern-.025em b}\kern-.08em
    T\kern-.1667em\lower.7ex\hbox{E}\kern-.125emX}}
\usepackage{subfiles}
\usepackage{pgfplotstable}
\usetikzlibrary{calc}
\newcommand\Min{-100}
\newcommand\Max{100}
\pgfplotsset{domain = \Min : \Max}
\pgfplotsset{compat=1.12, every axis/.append style={
		label style={font=\small},
		tick label style={font=\fontsize{3}{4}\selectfont} 
}}
\usepackage{mathtools}
\usepackage[caption=false]{subfig}
\usepackage{floatrow}
\usepackage{sidecap}
\usepackage{url}  
\usepackage{floatrow}
\usepackage{multirow}

\newcommand{\R}{\mathbb{R}}
\newcommand{\E}{\mathbb{E}}
\newcommand\norm[1]{\left\lVert#1\right\rVert}
\newcommand\independent{\protect\mathpalette{\protect\independenT}{\perp}}
\def\independenT#1#2{\mathrel{\rlap{$#1#2$}\mkern2mu{#1#2}}}

\usepackage{hyperref}
\newcommand{\email}[1]{\href{mailto:#1}{\nolinkurl{#1}}}

\providecommand{\keywords}[1]{\textbf{\textit{Index terms---}} #1}
\date{}

\usepackage{authblk}
	
\title{Generalization Comparison of Deep Neural Networks via Output Sensitivity
}
\author{Mahsa Forouzesh}
\author{Farnood Salehi}
\author{Patrick Thiran}
\affil{Information and Network Dynamics Group (INDY) \protect\\
	School of Computer and Communication Sciences (IC) \protect \\
	\'{E}cole Polytechnique F\'{e}d\'{e}rale de Lausanne \protect\\
	Switzerland \protect\\
\email{firstname.lastname@epfl.ch}}

\begin{document}
\maketitle

\begin{abstract}
	Although recent works have brought some insights into the performance improvement of techniques used in state-of-the-art deep-learning models, more work is needed to understand their generalization properties. We shed light on this matter by linking the loss function to the output's sensitivity to its input. We find a rather strong empirical relation between the output sensitivity and the variance in the bias-variance decomposition of the loss function, which hints on using sensitivity as a metric for comparing the generalization performance of networks, without requiring labeled data. We find that sensitivity is decreased by applying popular methods which improve the generalization performance of the model, such as (1) using a deep network rather than a wide one, (2) adding convolutional layers to baseline classifiers instead of adding fully-connected layers, (3) using batch normalization, dropout and max-pooling, and (4) applying parameter initialization techniques. 
\end{abstract}


	\section{Introduction}
	In machine-learning tasks, the main challenge a network designer faces is to find a model that learns the training data and that is able to predict the output of unseen data with high accuracy. 
	The first part is quite easily achievable by current over-parameterized deep neural networks, but the second part, referred to as generalization, demands careful expert hand-tuning \cite{lecun2015deep, goodfellow2016deep}. 
	Modern convolutional neural network (CNN) architectures that achieve state-of-the-art results in computer-vision tasks, such as ResNet~\cite{he2016deep} and VGG \cite{simonyan2014very}, attain high-generalization performance. Part of their success is due to recent advances in hardware and the availability of large amounts of data, but their generalization performance remains unequal. Therefore, knowing
	when and why some models generalize, still remain open questions to a large extent \cite{neyshabur2017exploring}.
	
	In this paper, by investigating the link between sensitivity and generalization, we get one step closer to understanding the generalization properties of deep neural networks. 
	Our findings suggest a relation between the sensitivity metric, a measure of uncertainty of the output with respect to input perturbations, and the variance term in the bias-variance decomposition of the test loss. This relation gives insight in the link between sensitivity and loss when the bias is small, 
	not only for classification tasks, but also for regression tasks. 
	
	Leveraging this relation, we can use the sensitivity metric to examine which network is more prone to overfitting. 
	Our numerical results suggest sensitivity as an appealing metric that captures the generalization improvements brought by a large class of architectures and techniques used in state-of-the-art models.
	In summary, we make the following contributions:
	\begin{itemize}
		\item We provide an approximate relation between sensitivity and generalization loss, via the relation between sensitivity and variance in the bias-variance decomposition of the loss. Our empirical results on state-of-the-art convolutional neural networks suggest a surprisingly strong match between experimental results and this (rather crude) approximation.
		\item We propose sensitivity as a promising architecture-selection metric and show that sensitivity, similarly to the test loss, promotes certain architectures compared to others. We in particular study the addition of convolutional layers versus fully-connected ones, and depth versus width. Sensitivity can potentially be used as a neural architecture search (NAS) tool, a priori (before training), to automate the architecture-design process.
		\item We provide an alternative explanation for the success of batch normalization in terms of sensitivity. We further give a new viewpoint on the performance improvement of dropout and max-pooling, as networks with these methods have a lower sensitivity alongside a lower generalization loss. We show that sensitivity retrieves the effectiveness of He and Xavier parameter initialization techniques.
	\end{itemize}

	\section{Related Work}
	
	To the best of our knowledge, \cite{dimopoulos1995use} was the first study to suggest a possible relation between sensitivity and generalization in multi-layer perceptrons, where the numerical results were limited to synthetic data.
	Recently, \cite{sokolic2017robust} suggested bounding the generalization error of deep neural networks with the spectral norm of the input-output Jacobian matrix, a measure of output sensitivity to its inputs. Reference \cite{novak2018sensitivity} empirically compares sensitivity, measured by the norm of the Jacobian of the \emph{output} of the softmax layer, and the generalization gap for fully-connected neural networks in image-classification tasks, leaving more complex architectures and other machine learning tasks as future work. 
	Our empirical results presented in Section~\ref{sec:bef_after}, together with the computations in Section~\ref{sec:bias_var}, suggest that sensitivity \emph{before} the softmax layer is related to the generalization loss, and that computing the sensitivity before (as in our paper) or after (as in \cite{novak2018sensitivity}) the softmax layer makes a strong difference (see e.g., Fig.~\ref{fig:softmax} in the appendix).
	In our work, we elaborate on the relation between sensitivity and loss for a wide range of settings, beyond fully-connected networks and image-classification tasks. We also show a rather strong match between the expression computed in Section~\ref{sec:bias_var} and the experimental results on state-of-the-art models.

	To avoid overfitting in deep-learning architectures, regularization techniques are applied, such as weight decay, early stopping, dropout \cite{srivastava2014dropout}, and batch normalization (BN) \cite{ioffe2015batch}. A popular explanation for the improved generalization of dropout is that it combines exponentially many networks to prevent overfitting \cite{srivastava2014dropout}. 
	Reference \cite{ioffe2015batch} argues that the reason for the success of BN is that it addresses the internal-covariant-shift phenomenon. However, \cite{santurkar2018does} argues against this belief and explains that the success of BN is due to its ability to make the optimization landscape smoother. In this paper, we look at the success of dropout and BN from another perspective: These methods decrease the output sensitivity to random input perturbations in a same manner as they decrease the test loss, resulting in better generalization performance.

	Designing neural network architectures is one of the main challenges in machine-learning tasks. One major line of work in this regard compares deep and shallow networks \cite{bengio2011expressive, mhaskar2017and, wu2019wider, ba2014deep, montufar2014number, simonyan2014very}.
	It is shown in \cite{telgarsky2016benefits} that to approximate a deep network, a shallow network requires an exponentially larger number of units per layer. 
	After finding a satisfactory architecture, the trainability of the network needs to be carefully assessed. To avoid exploding or vanishing gradients, \cite{glorot2010understanding} and \cite{he2015delving} introduce parameter initialization techniques that are widely used in current frameworks. By linking sensitivity and generalization, we present a new viewpoint on understanding the success of current state-of-the-art architectures and initialization techniques.

	Previous theoretical studies attempting to solve the mystery of generalization include generalization error (GE) bounds that use complexity measures such as VC-dimension and Rademacher complexities \cite{mohri2018foundations}. 
	Encouraged by the ability of neural networks to fit an entire randomly labeled dataset~\cite{zhang2016understanding},
	studies on data-dependent GE bounds have recently emerged~\cite{kawaguchi2017generalization,bartlett2017spectrally, arora2018stronger}. Computing a practical non-vacuous GE bound that completely captures the generalization properties of deep neural networks is still an evolving area of research~\cite{dziugaite2017computing, neyshabur2017exploring, nagarajan2019uniform}. In this paper, we do not study GE bounds. We propose sensitivity as a practical proxy for generalization in a large number of settings.

	There has been research on sensitivity analysis in neural networks with sigmoid and tanh activation functions \cite{dimopoulos1995use, fu1993sensitivity, zeng2001sensitivity}. Reference \cite{yang2013effective} introduces a sensitivity-based ensemble approach which selects individual networks with diverse sensitivity values from a pool of trained networks. Reference \cite{piche1995selection} performs a sensitivity analysis in neural networks to determine the required precision of the weight updates in each iteration. 
	In this work, we extend these results to networks with ReLU non-linearity with a different goal, which is to study the relation between sensitivity and generalization in state-of-the-art deep neural networks. Moreover, we provide a link between sensitivity and the variance in the bias-variance decomposition of the loss function.

	There have been recent attempts to predict the test loss for supervised-learning tasks \cite{novak2018sensitivity, jiang2018predicting, wang2018towards}. Reference \cite{chatterji2019intriguing} studies the module criticality, which is a weighted average over the distance of the network parameter vectors from their initial values. Although there seems to be a positive correlation between module criticality and generalization among different architectures, the correlation becomes negative when comparing the same architecture with different widths (as reported in Table~4 in \cite{chatterji2019intriguing}). 
	Reference~\cite{philipp2018nonlinearity} introduces the so-called non-linearity coefficient (NLC) as a gradient-based complexity measure of the neural network, which is empirically shown to be a predictor of the test error for fully-connected neural networks. 
	According to our results on both fully-connected and convolutional neural networks, sensitivity predicts the test loss, even before the networks are trained, which suggests sensitivity as a computationally inexpensive architecture-selection metric. 
	Among the mentioned metrics, the Jacobian norm, studied in~\cite{novak2018sensitivity}, does not require the computation of the parameter gradients nor the storage of large parameter vectors, as our metric, and therefore we compare it to sensitivity in Table~\ref{tab:jac}.

	\textbf{Paper Outline.} We formally define loss and sensitivity metrics in Section~\ref{sec:pre}. In Section \ref{sec:sen_loss}, we state the main findings of the paper and present the numerical and analytical results supporting them\footnote{The code to reproduce our experimental results is available at \url{https://github.com/mahf93/sensitivity}}. Later in Section~\ref{sec:explain}, we propose a possible proxy for generalization properties of certain structures and certain methods and present the empirical results for a regression task with the Boston housing dataset. Finally in Section~\ref{sec:dis}, we further discuss the observations followed up by a conclusion in Section~\ref{sec:conclusion}. The empirical results for image-classification tasks presented in the main part of the paper are on the CIFAR-10 dataset and the empirical results for MNIST and CIFAR-100 datasets are deferred to Appendix~\ref{app:mnist}.

	\section{Preliminaries}\label{sec:pre}
	
	Consider a supervised-learning task, where the model predicts a ground-truth output $y \in \mathcal{Y} \coloneqq \R^K$  for an input~$x \in \mathcal{X} \coloneqq \R^D$. The predictor $F_{\theta} : \mathcal{X} \rightarrow \mathcal{Y}$  is a deep neural network parameterized by the parameter vector~$\theta$ that is learned on the training dataset $\mathcal{D}_t$ by using the stochastic learning algorithm~$\mathcal{A}$. The training dataset~$\mathcal{D}_t$ and the testing (validation) dataset~$\mathcal{D}_v$ consist of i.i.d. samples drawn from the same data distribution $p$. With some abuse of notation, we use $\sim$ when the samples are uniformly drawn from a set of samples or from a probability distribution. 
	\subsection{Loss}\label{sec:loss_def}
	Our main focus is a classification task where the loss function is the cross-entropy criterion. The average test loss can be defined as
	\begin{align}\label{eq:cetestloss}
	L = \E_{\theta^*}[L_{\theta^*}] 
	= \E_{\theta^*}\left[\E_{(x,y) \sim \mathcal{D}_v} \left[- \sum_{k=1}^{K} y^k \log{F_{\theta^*}^k(x)}\right] \right] ,
	\end{align}
	where $\theta^*$ 
	is the random\footnote{The randomness is introduced by the stochastic optimization algorithm $\mathcal{A}$ and the randomized parameter initialization technique. } parameter vector found by $\mathcal{A}$, which minimizes the training loss defined on $\mathcal{D}_t$; $K$ is the number of classes and $F_{\theta^*}^k$ is the $k$-th entry of the vector~$F_{\theta^*}$, which is the output of the softmax layer, i.e.,~$F_{\theta^*}(x)=\textrm{softmax}(f_{\theta^*}(x))$, where~$f_{\theta^*}(x)$ is the output of the last layer of the network. In classification tasks, the output space is~$\mathcal{Y} \coloneqq [0,1]^K$, and the output vector is the probability assigned to each class.
	
	\subsection{Sensitivity}\label{sec:sen_def}
	Let us inject an external noise to the input of the network and compute the resulting noise in the output. If the original input vector is $x \in \mathcal{X}$ to which we add an i.i.d. normal noise vector ${\varepsilon_x \sim \mathcal{N}(0,\sigma^2_{\varepsilon_x} I)}$, then the output noise due to ${\varepsilon_x \in \mathcal{X}}$ is ${\varepsilon_y = f_{\theta}(x+\varepsilon_x) - f_{\theta}(x)}$. We use the variance of the output noise%
	, averaged over its $K$ entries, as a measure of sensitivity: ${S_\theta = \mathrm{Var}(\overline{\varepsilon_y})}$. The average sensitivity is therefore
	\begin{align}\label{eq:sen}
	S = \E_{\theta}[S_{\theta}] = \E_{\theta}\left[ \mathrm{Var}_{x, \varepsilon_x}\left[\frac{1}{K} \sum_{k=1}^{K} \varepsilon_y^k\right]\right] ,
	\end{align}
	where $\varepsilon_y^k$ is the $k$-th entry of the vector $\varepsilon_y$. 
	To distinguish the sensitivity $S$ computed on untrained networks from trained ones, we denote $S_{\textrm{before}} = \E_{\theta}[S_{\theta}]$ and~${S_{\textrm{after}} = \E_{\theta^*}[S_{\theta^*}]}$ when the expectation is over the network parameters before and after training, respectively. 
	We consider an ``unspecific`` sensitivity (meaning that the average is taken over all the entries of the output noise), which requires unlabeled data samples, as opposed to the ``specific`` sensitivity (which is limited to the output of the desired class) defined in~\cite{tartaglione2018learning}. %
	In this work, the input vectors $x$ used for computing $S$ are drawn from $\mathcal{D}_v$, so that given a new test data point, the sensitivity~$S$ predicts which trained network performs better for this particular point, and therefore gives a real-time uncertainty metric for predicting unseen data. For a few network architectures, we computed $S$ on the training set~$\mathcal{D}_t$ and observed that its value is practically the same as $S$ computed on the testing set~$\mathcal{D}_v$ (see Fig.~\ref{fig:test_vs_train} in the appendix), which suggests that $S$ as a generalization metric does not require sacrificing a set of training points for validation.

	\begin{figure*}[h]
		
	\begin{tikzpicture}[every mark/.append style={mark size=1.5pt}]
	
	\begin{axis}[width=7cm, height=4.5cm, legend pos=outer north east, legend columns=2, legend style={font=\small}, xlabel= {$\log(S_{\textrm{after}})$}, ylabel= { $\log(L)$}]
	
	\spaceskip0pt
	
	\addplot [red, only marks]  table [x=logm, y=logloss]   {\tableone};
	\addplot [blue, only marks]  table [x=logm, y=logloss]   {\tabletwo};
	\addplot [teal, only marks]  table [x=logm, y=logloss]   {\tablethree};
	\addplot [magenta, only marks, mark=triangle*]  table [x=logm, y=logloss]   {\tablefive};
	\addplot [green, only marks, mark=triangle*]  table [x=logm, y=logloss]   {\tablesix};
	\addplot [black, only marks, mark=o]  table [x=logm, y=logloss]   {\tableseven};
	\addplot [cyan, only marks, mark=o]  table [x=logm, y=logloss]   {\tableeight};
	
	\addplot [olive, solid, domain=-5:40]{0.5*(ln(0.5*0.9*0.2648/0.01))+0.5*x};

	\legend{Alexnet, Alexnet standard normal, 4 layer CNN standard normal, VGG16, VGG13 standard normal,ResNet18, ResNet18 standard normal, Equation (\ref{eq:main})}

	\node[scale=0.8] at (axis cs: 1,18) {$\rho=0.9707$};

	\end{axis}

	\end{tikzpicture}
	
	\caption{Test loss $L$ versus sensitivity $S_\text{after}$  for popular CNN architectures. The parameter initialization is Xavier \cite{glorot2010understanding} with uniform distribution, unless stated as standard normal distribution. The networks are trained on a subset of the CIFAR10 training dataset and are evaluated on the entire CIFAR10 test dataset. Each point of the plot indicates a network with a different number of channels and hidden units, and its coordinates~$\log(L)$ and $\log(S_\text{after})$ are averaged over 10 runs. For more details on configurations refer to Appendix \ref{app:exp}. The Pearson correlation coefficient $\rho$ between the data points is~$0.9707$. }
	\label{fig:state}
	
\end{figure*}
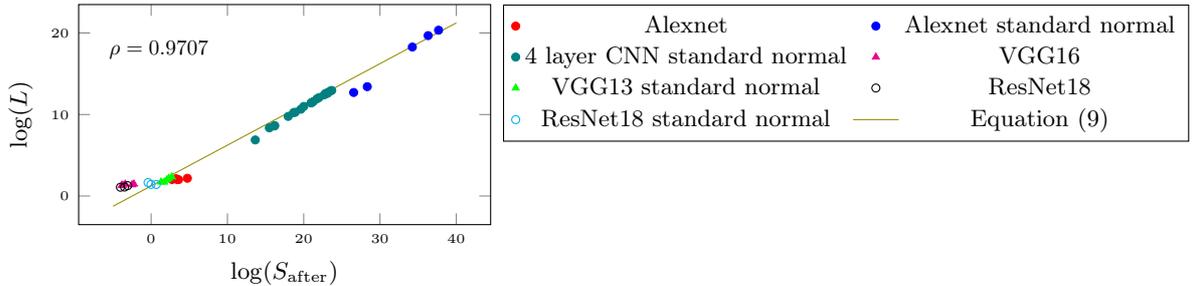

	\section{Sensitivity Versus Loss}\label{sec:sen_loss}
	\subsection{Numerical Experiments}
	
	An ideal predictor should 
	be robust
	: given similar inputs, the outputs should be close to each other.
	Assuming that the unseen data is drawn from the same distribution as the training data, the two concepts of robustness and generalization should therefore be related. Robustness here is the average-case robustness, not the worst-case robustness (adversarial robustness).
	We measure it by computing $S$ (Equation~(\ref{eq:sen})), and considering near-zero training loss, we refer to the test loss $L$ (Equation~(\ref{eq:cetestloss})) as the generalization error. 
	According to our observations on a wide set of experiments, including ResNets~\cite{he2016deep} and VGGs \cite{simonyan2014very}, we find a rather strong relation between $S$ and $L$. 
	State-of-the-art networks decrease the generalization error alongside with the sensitivity of the output of the network with respect to the input (Fig.~\ref{fig:state}).

	Many factors influence the generalization performance of deep-learning models, among which network topology, initialization technique, and regularization method. In Section~\ref{sec:explain}, we study the influence of each of these three factors on $S$ and keep all the other factors, including the learning algorithm, the same throughout the experiments. These experiments suggest 
	the use of $S_\text{after}$ as a proxy to the test loss, which is particularly advantageous for settings where labeled training data is limited; assessing generalization performance can then be done without having to sacrifice training data for the validation set. Furthermore, $S_\text{before}$ can potentially be used as an architecture-selection metric before training the models. We refer to fully-connected neural networks as FC, and to convolutional neural networks as CNN.

	\subsection{Bias-Variance Decomposition}\label{sec:bias_var}

	In this section, a crude approximate relation between sensitivity and generalization error is established through the link between sensitivity and the variance term in the bias-variance decomposition of the mean square error (MSE). First, we find the link between the cross-entropy loss and MSE. Next, we develop the relation between sensitivity and the variance term, and finally, the link between sensitivity $S$ and generalization error $L$.
	
	When the predictor $F_{\theta^*}(x)$ assigns the probability $F_{\theta^*}^c(x)$ to the correct class $c$ and $1-F_{\theta^*}^c(x)$ to another class (see Appendix \ref{app:prop2} for details), the cross-entropy loss $L$ can be approximated as\footnote{This is more accurate for over-confident predictors (see Appendix~\ref{app:prop2}).}
	\begin{equation}\label{eq:cemse}
	L \approx \E_{(x,y, \theta^*)} \left[\frac{1}{\sqrt{2}} \sqrt{\sum_{k=1}^{K} \left(F_{\theta^*}^k (x) - y^k \right)^2} \right] . 
	\end{equation}
	We roughly approximate the right-hand side in~(\ref{eq:cemse}) with $\sqrt{L_\text{MSE}/2}$, where $L_\text{MSE}$ is the mean square error criterion defined as
	\begin{align}\label{eq:msetestloss}
	L_{\text{MSE}} = \E_{\theta^*}[L_{{\theta^*}_\textrm{MSE}}]
	= \E_{\theta^*} \left[\E_{(x,y) \sim \mathcal{D}_v} \left[\norm{F_{\theta^*}(x)-y}^2_2\right] \right] .
	\end{align}
	Consider the classic notion of bias-variance decomposition for the MSE loss \cite{geman1992neural, tibshirani1996bias, neal2018modern, mehta2019high}, where the generalization error is the sum of three terms: bias, variance and noise, i.e., ${L_{\textrm{MSE}} = \varepsilon_{\textrm{bias}} + \varepsilon_{\textrm{variance}} + \varepsilon_{\textrm{noise}}}$. In this work, we consider the labels to be noiseless and neglect the third term~$\varepsilon_{\textrm{noise}}$. %
	The bias term is formally defined as
	\begin{equation}\label{eq:bias}
	\varepsilon_{\textrm{bias}} = \E_{x,y} \left[\norm{ \E_{\theta^*} [F_{\theta^*}(x)] - y}^2_2 \right],
	\end{equation}
	and the variance term is 
	\begin{equation}{\label{eq:var}}
	\varepsilon_{\textrm{variance}} = \sum_{k=1}^{K} \E_x \left[ \textrm{Var}_{\theta^*}(F_{\theta^*}^k(x))\right] .
	\end{equation}

	Let us now draw an again crude approximate relation between $\varepsilon_{\textrm{variance}}$ and $S$ under strong assumptions on the probability distributions of $\theta$, $x$, and $\varepsilon_x$ (refer to Appendix~\ref{app:prop1} for more details).
	Given a feed-forward neural network with $M$ hidden layers and $H_l$ units per layer, $1\leq l \leq M$, 
	where the non-linear activation function is positive homogeneous\footnote{ReLU is a positive homogeneous function with $\alpha=1$ and $\beta=0$.} with parameters $\alpha$ and $\beta$ (Equation~(\ref{eq:nonl}) in Appendix~\ref{app:prop1}), we have
	\begin{equation}\label{eq:prop1}
	\varepsilon_{\textrm{variance}} \approx \left(\frac{K-1}{K}\right) \left(S \cdot \frac{\sigma^{2}_{{x}}}{\sigma^{2}_{{\varepsilon_x}}}  + \Sigma \right)  ,
	\end{equation}
	where 
	\begin{equation}\label{eq:chi}
	\Sigma = \frac{1}{K} \sum_{l=1}^{M} \sigma^{2}_{{b_l}} \prod_{i=l+1}^{M} \left(\frac{\alpha^2+\beta^2}{2}\right) \sigma^{2}_{{w_{i}}} H_{i} ,
	\end{equation}
	where $K$ is the number of units in the output of the softmax layer and $\sigma^{2}_{{w_l}}$, $\sigma^{2}_{{b_l}}$, $\sigma^{2}_{{x}}$, and $\sigma^{2}_{{\varepsilon_x}}$ are the second moment of weights and biases of layer $l$, input $x$ and input noise~$\varepsilon_x$, respectively. Equation (\ref{eq:chi}) can be extended to convolutional neural networks\footnote{$fan_{in}=$ the number of input channels $*$ the kernel size} by replacing $H_i$ with $fan_{in}$ of layer $i$. %
	
	Given an infinite amount of training data, the bias represents the best performance of the model, which can be approximated by the training loss \cite{mehta2019high, andrewbiasvar, biasvar}. In deep learning settings (and thus in our experiments), the training loss is close to zero, hence if we neglect the bias term $\varepsilon_{\textrm{bias}}$ in the decomposition of $L_{\textrm{MSE}}$ we have
	\begin{equation}\label{eq:main}
		L \approx \sqrt{\frac{1}{2} \left( \frac{K-1}{K}\right) \left( S \cdot \frac{\sigma^{2}_{{x}}}{\sigma^{2}_{{\varepsilon_x}}} + \Sigma \right)} ,
	\end{equation}
	where $\Sigma$ is given by~(\ref{eq:chi}). In the experiments, we observe that $\sigma_{b_l}^2$ is usually very small or zero (for instance in ResNets because $b_l=0$), making $\Sigma \approx 0$.
	
	According to~(\ref{eq:prop1}) and the relation between $L_\text{MSE}$ and $L$, 
	to compare networks with a small value of $\varepsilon_\textrm{bias}$ (which is usually the case in deep neural networks where the bias is approximated by the near-zero training loss), the test loss can be approximated using the sensitivity by~(\ref{eq:main}). Despite the strong assumptions and crude approximations to get~(\ref{eq:main}), the numerical experiments show a rather surprisingly good match with~(\ref{eq:main}) (Figs.~\ref{fig:state}, \ref{fig:fcnn} and \ref{fig:init}), even if $\Sigma$ is neglected in~(\ref{eq:main}). It is interesting to note that the right-hand side of~(\ref{eq:main}) is computed without requiring labeled data points, whereas the left-hand side requires the ground-truth output vector $y$.
	 
	If $\varepsilon_{\textrm{bias}}$ can no longer be approximated by the training loss, which may in part explain the poorer match in lower values of $S_\text{after}$ in Fig.~\ref{fig:state}, we need more training data to make this approximation valid. In Section~\ref{app:bias} we train the networks with more data samples and observe that numerical results become closer to~(\ref{eq:main}).

	\subsection{Sensitivity Before Versus After The Softmax Layer}\label{sec:bef_after}
	\begin{table}[h]
		\caption{Pearson's correlation coefficient $\rho$ between each metric ($S$, $J$) and the test loss ($L$), and average computation time (in seconds) of each metric for VGG13, VGG16, ResNet18 and ResNet34 networks trained on the CIFAR-10 dataset. The test accuracy of the networks are up to $87\%$.} 
		\begin{center}
		\begin{tabular}{|c|c|c|}
			\hline
			Metric & $\rho$ & Computation time \\
			\hline
			$J$ after softmax & 0.116 & 1.166 $\pm$ 0.111          \\
			\hline
			$J$ before softmax & 0.414                &       1.165    $\pm$ 0.111        \\
			\hline
			$S$ after softmax & 0.381 & 0.086 $\pm$ 0.006\\
			\hline
			 $S$ before softmax  & \textbf{0.648}      & \textbf{0.085} $\pm$ 0.006         \\
			\hline
		\end{tabular}
	\label{tab:jac}
	\end{center}
	\end{table}

	It is interesting to compare the sensitivity $S$ given by~(\ref{eq:sen}) with the Frobenius norm of the Jacobian matrix~$J$ of the output of the softmax layer~\cite{novak2018sensitivity}, in terms of their ability to gauge the generalization error $L$.  
	A practical motivation for using $S$ instead of $J$ in real-world applications is computational tractability: to find the network architecture(s) with the best generalization ability among a collection of trained networks, the computation of $S$ does not require to make a backward pass through each network architecture, contrary to$J$. In Table~\ref{tab:jac} we observe that computing Jacobian is more than 10 times slower than computing sensitivity.
	But the main motivation for using $S$ is that it is computed \emph{before} and not \emph{after} the softmax layer, contrary to $J$ in \cite{novak2018sensitivity}. Because of the chain rule, $J$ depends on the derivative of the softmax function with respect to the logits, which has very low values for highly confident predictors (the ones which assign a very high probability to one class and almost zero probability to the other classes). For instance, if the predictor erroneously assigns a high probability to a wrong class, the derivative of the softmax function is very low, resulting in a very low $J$. In this case, $J$ would be misleading as it would mistakenly indicate good generalization. In contrast, $S$ does not depend on the confidence level of the predictor. The difference is illustrated in Table~\ref{tab:jac} (see also Fig.~\ref{fig:softmax} in the appendix), where the strong correlation between $S$ and $L$ (and the good match with~(\ref{eq:main})) is not found when $S$ is replaced by the sensitivity \emph{after} the softmax layer. Therefore, we observe from Table~\ref{tab:jac} that $S$ computed before the softmax layer (given by~(\ref{eq:sen})), is preferred to $J$ (defined in~\cite{novak2018sensitivity}), both in terms of correlation to the test loss and of computation time.

	\section{Sensitivity as a Proxy for Generalization}\label{sec:explain}
	
	\begin{figure*}[h]
		\sidesubfloat[]{
			\scalebox{0.45}{
				\begin{tikzpicture}[every mark/.append style={mark size=3pt}]
				
				\begin{axis}[name=boundary, width=13cm, height=8cm, legend pos=north west, legend style={fill=none, font=\Large}, xlabel= {$\log(S_\text{after})$}, ylabel= { $\log(L)$}, every axis/.append style={label style={font=\LARGE},
					tick label style={font=\large} }, xtick distance=5, ytick distance=5]
				
				\spaceskip0pt
				
				\addplot [stack plots=x, fill=none, draw=none, forget plot]   table [x=logl, y=logloss]   {\tableone} \closedcycle; 
				\addplot [stack plots=x, fill=teal!50, opacity=0.4, draw opacity=0, area legend, forget plot]   table [x expr=\thisrow{logu}-\thisrow{logl}, y = logloss]   {\tableone} \closedcycle;
				\addplot [stack plots=x, stack dir=minus, forget plot, draw=none] table [x=logu, y=logloss] {\tableone};
				
				\addplot [stack plots=x, fill=none, draw=none, forget plot]   table [x=logl, y=logloss]   {\tabletwo} \closedcycle; 
				\addplot [stack plots=x, fill=blue!50, opacity=0.4, draw opacity=0, area legend, forget plot]   table [x expr=\thisrow{logu}-\thisrow{logl}, y = logloss]   {\tabletwo} \closedcycle;
				\addplot [stack plots=x, stack dir=minus, forget plot, draw=none] table [x=logu, y=logloss] {\tabletwo};
				
				\addplot [stack plots=x, fill=none, draw=none, forget plot]   table [x=logl, y=logloss]   {\tableonec} \closedcycle; 
				\addplot [stack plots=x, fill=magenta!50, opacity=0.4, draw opacity=0, area legend, forget plot]   table [x expr=\thisrow{logu}-\thisrow{logl}, y = logloss]   {\tableonec} \closedcycle;
				\addplot [stack plots=x, stack dir=minus, forget plot, draw=none] table [x=logu, y=logloss] {\tableonec};

				\addplot [teal, only marks]  table [x=logm, y=logloss]   {\tableone};\label{fc2}
				\addplot [blue, only marks]  table [x=logm, y=logloss]   {\tabletwo};\label{fc3}
				\addplot [magenta, only marks]  table [x=logm, y=logloss]   {\tableonec};\label{conv1}
				
				\node[scale=1.5] at (axis cs: 10,4.5) {$\rho=0.9739$};

				\filldraw [fill=green!50, draw=black, thick] (axis cs:7.15, 4.65) circle [radius=13pt]; \label{m1}
				\filldraw [fill=yellow, draw=black, thick] (axis cs:8.67,5.63) circle [radius=13pt]; \label{m2}
				
				\legend{2 layer FC (2 fc layers), 3 layer FC (2 fc layers + 1 fc layer), 1 layer CNN (2 fc layers + 1 conv layer)}

				\filldraw [fill=white, draw=black, thick] (axis cs:-0.1, 5.28) circle [radius=11pt];
				\node[draw,inner sep=0pt,above left=0.5em] at (2.35,4.5) {\Large
					\begin{tabular}{cccl}
					& &  \\
					& &  mark\\
					& &  \\
					\end{tabular}};
				\end{axis}

				\end{tikzpicture}
				
		}}
		\sidesubfloat[]{
			\scalebox{0.45}{
		\begin{tikzpicture}[every mark/.append style={mark size=3pt}]
		
		\begin{axis}[width=13cm, height=8cm, legend pos=north west, legend style={fill=none, font=\Large}, xlabel= {$\log(S_{\text{after}})$}, ylabel= { $\log(L)$}, every axis/.append style={label style={font=\LARGE},
			tick label style={font=\large} }, xtick distance=10, ytick distance=5]
		\addplot [stack plots=x, fill=none, draw=none, forget plot]   table [x=logl, y=logloss]   {\tableone} \closedcycle;
		\addplot [stack plots=x, fill=teal!50, opacity=0.4, draw opacity=0, area legend, forget plot]   table [x expr=\thisrow{logu}-\thisrow{logl}, y=logloss]   {\tableone} \closedcycle;
		\addplot [stack plots=x, stack dir=minus, forget plot, draw=none] table [x=logu, y=logloss] {\tableone};
		
		\addplot [stack plots=x, fill=none, draw=none, forget plot]   table [x=logl, y=logloss]   {\tabletwo} \closedcycle;
		\addplot [stack plots=x, fill=blue!50, opacity=0.4, draw opacity=0, area legend, forget plot]   table [x expr=\thisrow{logu}-\thisrow{logl}, y=logloss]   {\tabletwo} \closedcycle;
		\addplot [stack plots=x, stack dir=minus, forget plot, draw=none] table [x=logu, y=logloss] {\tabletwo};
		
		\addplot [stack plots=x, fill=none, draw=none, forget plot]   table [x=logl, y=logloss]   {\tablethree} \closedcycle;
		\addplot [stack plots=x, fill=red!50, opacity=0.4, draw opacity=0, area legend, forget plot]   table [x expr=\thisrow{logu}-\thisrow{logl}, y = logloss]   {\tablethree} \closedcycle;
		\addplot [stack plots=x, stack dir=minus, forget plot, draw=none] table [x=logu, y=logloss] {\tablethree};
		
		\addplot [stack plots=x, fill=none, draw=none, forget plot]   table [x=logl, y=logloss]   {\tablefour} \closedcycle;
		\addplot [stack plots=x, fill=cyan!50, opacity=0.4, draw opacity=0, area legend, forget plot]   table [x expr=\thisrow{logu}-\thisrow{logl}, y=logloss]   {\tablefour} \closedcycle;
		\addplot [stack plots=x, stack dir=minus, forget plot, draw=none] table [x=logu, y=logloss] {\tablefour};
		
		\addplot [stack plots=x, fill=none, draw=none, forget plot]   table [x=logl, y=logloss]   {\tablefive} \closedcycle;
		\addplot [stack plots=x, fill=magenta!50, opacity=0.4, draw opacity=0, area legend, forget plot]   table [x expr=\thisrow{logu}-\thisrow{logl}, y=logloss]   {\tablefive} \closedcycle;
		\addplot [stack plots=x, stack dir=minus, forget plot, draw=none] table [x=logu, y=logloss] {\tablefive};
		
		\addplot [stack plots=x, fill=none, draw=none, forget plot]   table [x=logl, y=logloss]   {\tablesix} \closedcycle;
		\addplot [stack plots=x, fill=black!50, opacity=0.4, draw opacity=0, area legend, forget plot]   table [x expr=\thisrow{logu}-\thisrow{logl}, y=logloss]   {\tablesix} \closedcycle;
		\addplot [stack plots=x, stack dir=minus, forget plot, draw=none] table [x=logu, y=logloss] {\tablesix};
		
		\addplot [teal, only marks]  table [x=logm, y=logloss]   {\tableone};
		\addplot [blue, only marks]  table [x=logm, y=logloss]   {\tabletwo};
		\addplot [red, only marks]  table [x=logm, y=logloss]   {\tablethree};
		\addplot [cyan, only marks]  table [x=logm, y=logloss]   {\tablefour};
		\addplot [magenta, only marks]  table [x=logm, y=logloss]   {\tablefive};
		\addplot [black, only marks]  table [x=logm, y=logloss]   {\tablesix};
		
		\addplot [olive, solid, domain=3:35]{0.5*(ln(0.5*0.9*0.2648/0.01))+0.5*x};
		
		\legend{2 layer FC, 3 layer FC, 4 layer FC, 5 layer FC, 6 layer FC, 7 layer FC, Equation (\ref{eq:main})}
		
		
		\node[scale=1.5] at (axis cs: 30,4) {$\rho=0.9625$};
		
		\filldraw [fill=yellow, draw=black, thick] (axis cs:16.98,9.60) circle [radius=13pt];
		
		\filldraw [fill=white, draw=black, thick] (axis cs:13.6, 16.7) circle [radius=11pt];
		\node[draw,inner sep=0pt,above left=0.5em] at (20.95,13.5) {\Large
			\begin{tabular}{cccl}
			& &  \\
			& &  mark\\
			& &  \\
			\end{tabular}};
		
		\end{axis}
		
		\end{tikzpicture}
	}
	}\\
\sidesubfloat[]{
	\scalebox{0.45}{
		\begin{tikzpicture}[every mark/.append style={mark size=3pt}]
		
		\begin{axis}[width=10.5cm, height=8cm, legend pos=north west, legend style={fill=none, font=\Large}, xlabel= {$\log(S_{\text{after}})$}, ylabel= { $\log(L)$}, every axis/.append style={label style={font=\LARGE},
			tick label style={font=\large} }, xtick distance=10, ytick distance=5]
		
		\addplot [stack plots=x, fill=none, draw=none, forget plot]   table [x=logl, y=logloss]   {\tableonec} \closedcycle;
		\addplot [stack plots=x, fill=red!50, opacity=0.4, draw opacity=0, area legend, forget plot]   table [x expr=\thisrow{logu}-\thisrow{logl}, y=logloss]   {\tableonec} \closedcycle;
		\addplot [stack plots=x, stack dir=minus, forget plot, draw=none] table [x=logl, y=logloss] {\tableonec};
		
		\addplot [stack plots=x, fill=none, draw=none, forget plot]   table [x=logl, y=logloss]   {\tabletwoc} \closedcycle;
		\addplot [stack plots=x, fill=blue!50, opacity=0.4, draw opacity=0, area legend, forget plot]   table [x expr=\thisrow{logu}-\thisrow{logl}, y=logloss]   {\tabletwoc} \closedcycle;
		\addplot [stack plots=x, stack dir=minus, forget plot, draw=none] table [x=logu, y=logloss] {\tabletwoc};
		
		\addplot [stack plots=x, fill=none, draw=none, forget plot]   table [x=logl, y=logloss]   {\tablethreec} \closedcycle;
		\addplot [stack plots=x, fill=teal!50, opacity=0.4, draw opacity=0, area legend, forget plot]   table [x expr=\thisrow{logu}-\thisrow{logl}, y=logloss]   {\tablethreec} \closedcycle;
		\addplot [stack plots=x, stack dir=minus, forget plot, draw=none] table [x=logu, y=logloss] {\tablethreec};
		
		\addplot [stack plots=x, fill=none, draw=none, forget plot]   table [x=logl, y=logloss]   {\tablefourc} \closedcycle;
		\addplot [stack plots=x, fill=magenta!50, opacity=0.4, draw opacity=0, area legend, forget plot]   table [x expr=\thisrow{logu}-\thisrow{logl}, y=logloss]   {\tablefourc} \closedcycle;
		\addplot [stack plots=x, stack dir=minus, forget plot, draw=none] table [x=logu, y=logloss] {\tablefourc};
		
		\addplot [red, only marks]  table [x=logm, y=logloss]   {\tableonec};
		\addplot [blue, only marks]  table [x=logm, y=logloss]   {\tabletwoc};
		\addplot [teal, only marks]  table [x=logm, y=logloss]   {\tablethreec};
		\addplot [magenta, only marks]  table [x=logm, y=logloss]   {\tablefourc};
		\addplot [olive, solid, domain=4:24]{0.5*(ln(0.5*0.9*0.2648/0.01))+0.5*x};

		\legend{1 layer CNN, 2 layer CNN, 3 layer CNN, 4 layer CNN, Equation (\ref{eq:main})}
		
		\node[scale=1.5] at (axis cs: 21,4) {$\rho=0.9658$};
		
		\end{axis}
		
		\end{tikzpicture}}
}
		\sidesubfloat[]{
			\scalebox{0.45}{
		\begin{tikzpicture}[every mark/.append style={mark size=3pt}]
		
		
		\begin{axis}[width=10.5cm, height=8cm, legend pos=north west, legend style={fill=none, font=\Large}, xlabel= {$\log(S_{\text{after}})$}, ylabel= { $\log(L)$}, every axis/.append style={label style={font=\LARGE},
			tick label style={font=\large} }, xtick distance=5, ytick distance=5]{
			\addplot [stack plots=x, fill=none, draw=none, forget plot]   table [x=logl, y=logloss]   {\tablethree} \closedcycle;
			\addplot [stack plots=x, fill=teal!50, opacity=0.4, draw opacity=0, area legend, forget plot]   table [x expr=\thisrow{logu}-\thisrow{logl}, y=logloss]   {\tablethree} \closedcycle;
			\addplot [stack plots=x, stack dir=minus, forget plot, draw=none] table [x=logu, y=logloss] {\tablethree};
			
			\addplot [stack plots=x, fill=none, draw=none, forget plot]   table [x=logl, y=logloss]   {\tablethreed} \closedcycle;
			\addplot [stack plots=x, fill=magenta!50, opacity=0.4, draw opacity=0, area legend, forget plot]   table [x expr=\thisrow{logu}-\thisrow{logl}, y=logloss]   {\tablethreed} \closedcycle;
			\addplot [stack plots=x, stack dir=minus, forget plot, draw=none] table [x=logu, y=logloss] {\tablethreed};
			
			\addplot [stack plots=x, fill=none, draw=none, forget plot]   table [x=logl, y=logloss]   {\tablethreebn} \closedcycle;
			\addplot [stack plots=x, fill=black!50, opacity=0.4, draw opacity=0, area legend, forget plot]   table [x expr=\thisrow{logu}-\thisrow{logl}, y=logloss]   {\tablethreebn} \closedcycle;
			\addplot [stack plots=x, stack dir=minus, forget plot, draw=none] table [x=logu, y=logloss] {\tablethreebn};

			\addplot [teal, only marks]  table [x=logm, y=logloss]   {\tablethree};
			\addplot [magenta, only marks]  table [x=logm, y=logloss]   {\tablethreed};
			\addplot [black, only marks]  table [x=logm, y=logloss]   {\tablethreebn};
			\addplot [olive, solid, domain=0:17]{0.5*(ln(0.5*0.9*0.2648/0.01))+0.5*x};
			
			\legend{4 layer FC, 4 layer FC+dropout, 4 layer FC+BN, Equation (\ref{eq:main})}}

			\node[scale=1.5] at (axis cs: 14,2) {$\rho=0.9127$};
			
		\end{axis}

		\end{tikzpicture}
			
		}}
	\sidesubfloat[]{
		\scalebox{0.45}{
		\begin{tikzpicture}[every mark/.append style={mark size=3pt}]
		
		\begin{axis}[width=10.5cm, height=8cm, legend pos=north west, legend style={fill=none, font=\Large}, xlabel= {$\log(S_\text{after})$}, ylabel= {$\log(L)$}, every axis/.append style={label style={font=\LARGE},
			tick label style={font=\large} }, xtick distance=10, ytick distance=5]{
			
			\addplot [stack plots=x, fill=none, draw=none, forget plot]   table [x=logl, y=logloss]   {\tablethreec} \closedcycle;
			\addplot [stack plots=x, fill=teal!50, opacity=0.4, draw opacity=0, area legend, forget plot]   table [x expr=\thisrow{logu}-\thisrow{logl}, y=logloss]   {\tablethreec} \closedcycle;
			\addplot [stack plots=x, stack dir=minus, forget plot, draw=none] table [x=logu, y=logloss] {\tablethreec};
			
			\addplot [stack plots=x, fill=none, draw=none, forget plot]   table [x=logl, y=logloss]   {\tablethreebnc} \closedcycle;
			\addplot [stack plots=x, fill=black!50, opacity=0.4, draw opacity=0, area legend, forget plot]   table [x expr=\thisrow{logu}-\thisrow{logl}, y=logloss]   {\tablethreebnc} \closedcycle;
			\addplot [stack plots=x, stack dir=minus, forget plot, draw=none] table [x=logu, y=logloss] {\tablethreebnc};
			
			\addplot [stack plots=x, fill=none, draw=none, forget plot]   table [x=logl, y=logloss]   {\tablethreedc} \closedcycle;
			\addplot [stack plots=x, fill=magenta!50, opacity=0.4, draw opacity=0, area legend, forget plot]   table [x expr=\thisrow{logu}-\thisrow{logl}, y=logloss]   {\tablethreedc} \closedcycle;
			\addplot [stack plots=x, stack dir=minus, forget plot, draw=none] table [x=logu, y=logloss] {\tablethreedc};
			
			\addplot [stack plots=x, fill=none, draw=none, forget plot]   table [x=logl, y=logloss]   {\tablethreemaxc} \closedcycle;
			\addplot [stack plots=x, fill=lime!50, opacity=0.4, draw opacity=0, area legend, forget plot]   table [x expr=\thisrow{logu}-\thisrow{logl}, y=logloss]   {\tablethreemaxc} \closedcycle;
			\addplot [stack plots=x, stack dir=minus, forget plot, draw=none] table [x=logu, y=logloss] {\tablethreemaxc};
			
			\addplot [teal, only marks]  table [x=logm, y=logloss]   {\tablethreec};
			\addplot [black, only marks]  table [x=logm, y=logloss]   {\tablethreebnc};
			\addplot [magenta, only marks]  table [x=logm, y=logloss]   {\tablethreedc};
			\addplot [lime, only marks]  table [x=logm, y=logloss]   {\tablethreemaxc};
			\addplot [olive, solid, domain=0:22]{0.5*(ln(0.5*0.9*0.2648/0.01))+0.5*x};

			\legend{3 layer CNN, 3 layer CNN+BN, 3 layer CNN+dropout, 3 layer CNN+max-pooling, Equation (\ref{eq:main})}}
		
			\node[scale=1.5] at (axis cs: 17,2) {$\rho=0.9469$};
		\end{axis}
	
\end{tikzpicture}
}}
	\caption{Test loss $L$ versus sensitivity $S_\text{after}$ for networks trained on a subset of the CIFAR-10 training dataset where the network parameters are initially drawn from a standard normal distribution. Each point of the plot indicates a network with a different number of channels and hidden units, and its coordinates $\log(L)$ and~$\log(S_\text{after})$ are averaged over 10 runs. The shaded areas are contained by the minimum and maximum values of $\log(S_\text{after})$ over multiple runs of each experiment (point). \textbf{(a)} Comparison between adding a convolutional layer and adding a fully-connected layer to a baseline classifier that is a fully-connected neural network with one hidden layer. \textbf{(b)} Fully-connected neural networks. \textbf{(c)} Convolutional neural networks. \textbf{(d)} 4-layer FC trained with or without regularization. \textbf{(e)} 3-layer CNN trained with or without regularization. }
	\label{fig:fcnn}
	\end{figure*}
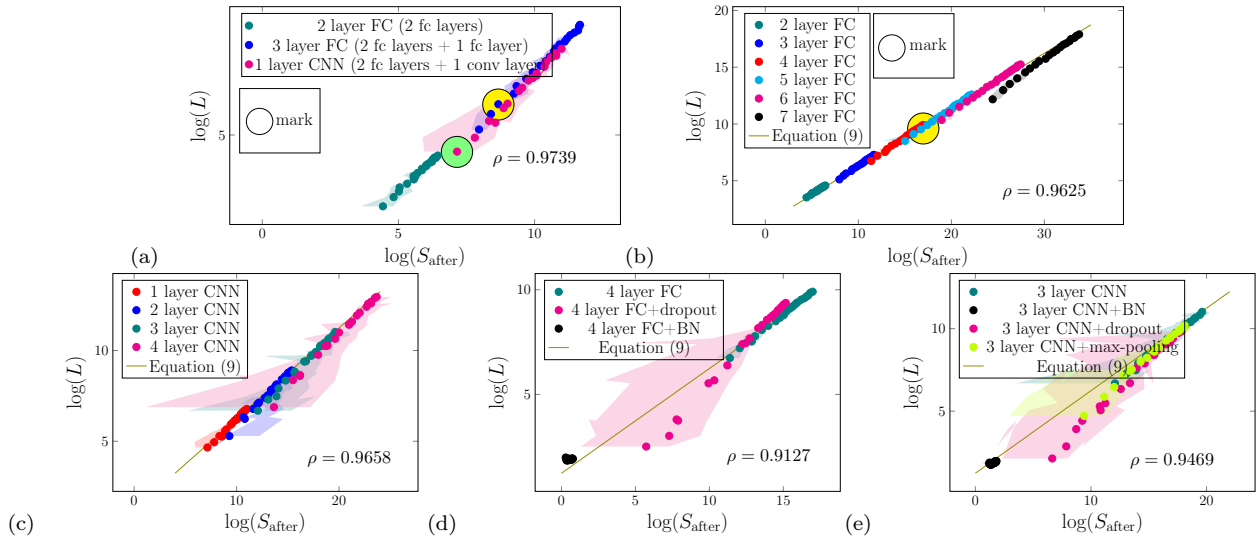

	In this section, %
	we argue that methods improving the generalization performance of neural networks remarkably reduce the sensitivity $S$. 
	We also present the experimental results for a regression task.

	\subsection{Comparing Different Architectures}\label{sec:compare}

	\textbf{Convolutional vs Fully-Connected Layers.}
	The relation between the sensitivity $S$ and the generalization error~$L$ supports the common view that CNNs outperform FCs in image-classification tasks. %
	In Fig.~\ref{fig:fcnn}~(a) we empirically observe that, given a CNN and an FC with the same number of parameters, the CNN has lower sensitivity and test loss than the FC. Moreover, some CNNs with more parameters than FCs have both lower sensitivity and lower test loss, even though they are more over-parameterized.\\
	Let us start from a baseline classifier with one hidden layer (2 layers in total displayed in teal blue points in Fig.~\ref{fig:fcnn}~(a), where each point represents a network with a different number of hidden units). We compare the effect of adding another fully-connected layer with adding a convolutional layer in Fig.~\ref{fig:fcnn}~(a). We vary the number of parameters of 1-layer CNNs (which consist of 2 fully-connected (fc) layers and 1 conv layer, displayed by pink points) from 450k to 10M by increasing the number of channels and hidden units, whereas the number of parameters for 3-layer FCs varies from 320k to 1.7M (displayed by dark blue points). Despite the large number of parameters of CNNs, they suffer from less overfitting and have a lower sensitivity $S$ than FCs.
	Next, let us compare a FC to a CNN with the same number of parameters in Fig.~\ref{fig:fcnn}~(a): A 3-layer FC with 140 units in each layer (yellow mark) and a 1-layer CNN with 5 channels and 100 units (green mark), both have 450k parameters. The CNN has remarkably lower sensitivity and test loss than the FC, which indicates better performance compared to the FC with the same number of parameters. 
	
	\textbf{Depth vs Width.} 
	Consider a feedforward FC with ReLU activation function %
	where all the network parameters follow the standard normal distribution and are independent from each other and from the input. If we have $M$ layers with $H$ units in each hidden layer, $K$ units in the output layer and $D$ units in the input layer, then (see Appendix~\ref{app:prop3} for details)
	\begin{equation}\label{eq:senfc}
		S = \frac{D}{K}\left(\frac{H}{2} \right)^M \sigma^2_{\varepsilon_x} .
	\end{equation}
	According to (\ref{eq:senfc}), considering two neural networks with the same value for $H^M$, one deep and narrow (higher $M$ and lower $H$), and the other one shallow and wide (lower $M$ and higher~$H$), 
	the deeper network has lower sensitivity $S$. Assuming that both networks have near-zero training losses, depth is better than width regarding generalization in fully-connected neural networks. The empirical results in Fig.~\ref{fig:fcnn}~(b) support~(\ref{eq:senfc}). For instance, a 4-layer FC with 500 units per layer (the top right most point among all 4-layer FCs, indicated by a yellow mark), has the same value for $({H}/{2})^M$ as a 5-layer FC with 165 units per layer (the 4th point among 5-layer FCs, which exactly matches the yellow mark). In Fig.~\ref{fig:fcnn}~(b), these two networks have the same values of both $S_\text{after}$ and $L$, and all narrower 5-layer networks (with 100, 120, and 140 units) have better performance than the wide 4-layer network (with 500 units). A similar trend is observed for CNNs in Fig.~\ref{fig:fcnn}~(c): having a narrower and deeper CNN is preferable to having a wider and shallower CNN.
	
	\subsection{Regularization Techniques}
	Figs.~\ref{fig:fcnn}~(d) and~(e) show the sensitivity $S_\text{after}$ versus the test loss $L$, for different regularization methods. In particular, we study the effect of dropout \cite{srivastava2014dropout} and batch normalization (BN) \cite{ioffe2015batch} on the sensitivity in the FCs; and we apply dropout, BN and max-pooling for the CNNs. The results are consistent with the relation between sensitivity $S_\text{after}$ and loss $L$. For all these regularization techniques, we observe a shift of the points towards the bottom left. This shift shows that these techniques known to improve generalization simultaneously decrease the network sensitivity to input perturbations. This is particularly noticeable in the BN case, where both the sensitivity and test loss decrease dramatically. This suggests that batch normalization improves performance by making the network less sensitive to input perturbations. 
	
	\subsection{Initialization Methods}
	
	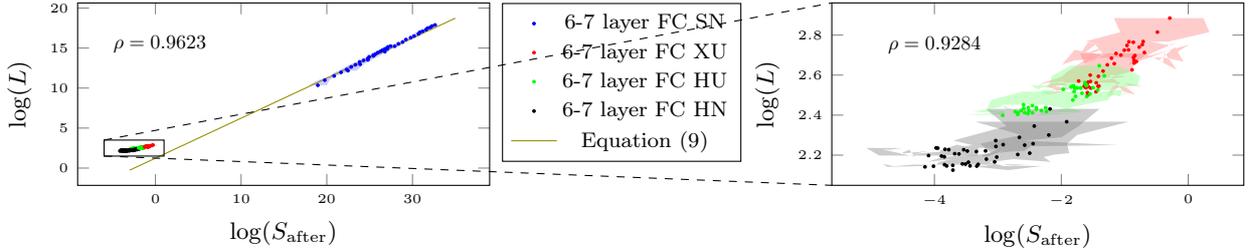
\begin{figure*}[t]
	\begin{tikzpicture}[every mark/.append style={mark size=0.5pt}]
	
	\begin{axis}[name=ax1, width=7cm, height=4cm, legend pos=outer north east,  legend style={fill=none, font=\footnotesize}, xlabel= {$\log(S_{\text{after}})$}, ylabel= { $\log(L)$}, every axis/.append style={label style={font=\small} }, xtick distance=10, ytick distance=5]

	\addplot [stack plots=x, fill=none, draw=none, forget plot]   table [x=logl, y=logloss]   {\tablefive} \closedcycle;
	\addplot [stack plots=x, fill=blue!50, opacity=0.4, draw opacity=0, area legend, forget plot]   table [x expr=\thisrow{logu}-\thisrow{logl}, y=logloss]   {\tablefive} \closedcycle;
	\addplot [stack plots=x, stack dir=minus, forget plot, draw=none] table [x=logu, y=logloss] {\tablefive};
	
	\addplot [stack plots=x, fill=none, draw=none, forget plot]   table [x=logl, y=logloss]   {\tablefivehen} \closedcycle;
	\addplot [stack plots=x, fill=black!50, opacity=0.4, draw opacity=0, area legend, forget plot]   table [x expr=\thisrow{logu}-\thisrow{logl}, y=logloss]   {\tablefivehen} \closedcycle;
	\addplot [stack plots=x, stack dir=minus, forget plot, draw=none] table [x=logu, y=logloss] {\tablefivehen};
	
	\addplot [stack plots=x, fill=none, draw=none, forget plot]   table [x=logl, y=logloss]   {\tablefiveheu} \closedcycle;
	\addplot [stack plots=x, fill=green!50, opacity=0.4, draw opacity=0, area legend, forget plot]   table [x expr=\thisrow{logu}-\thisrow{logl}, y=logloss]   {\tablefiveheu} \closedcycle;
	\addplot [stack plots=x, stack dir=minus, forget plot, draw=none] table [x=logu, y=logloss] {\tablefiveheu};
	
	\addplot [stack plots=x, fill=none, draw=none, forget plot]   table [x=logl, y=logloss]   {\tablefivexu} \closedcycle;
	\addplot [stack plots=x, fill=red!50, opacity=0.4, draw opacity=0, area legend, forget plot]   table [x expr=\thisrow{logu}-\thisrow{logl}, y=logloss]   {\tablefivexu} \closedcycle;
	\addplot [stack plots=x, stack dir=minus, forget plot, draw=none] table [x=logu, y=logloss] {\tablefivexu};
	
	\addplot [stack plots=x, fill=none, draw=none, forget plot]   table [x=logl, y=logloss]   {\tablesix} \closedcycle;
	\addplot [stack plots=x, fill=blue!50, opacity=0.4, draw opacity=0, area legend, forget plot]   table [x expr=\thisrow{logu}-\thisrow{logl}, y=logloss]   {\tablesix} \closedcycle;
	\addplot [stack plots=x, stack dir=minus, forget plot, draw=none] table [x=logu, y=logloss] {\tablesix};
	
	\addplot [stack plots=x, fill=none, draw=none, forget plot]   table [x=logl, y=logloss]   {\tablesixhen} \closedcycle;
	\addplot [stack plots=x, fill=black!50, opacity=0.4, draw opacity=0, area legend, forget plot]   table [x expr=\thisrow{logu}-\thisrow{logl}, y=logloss]   {\tablesixhen} \closedcycle;
	\addplot [stack plots=x, stack dir=minus, forget plot, draw=none] table [x=logu, y=logloss] {\tablesixhen};
	
	\addplot [stack plots=x, fill=none, draw=none, forget plot]   table [x=logl, y=logloss]   {\tablesixheu} \closedcycle;
	\addplot [stack plots=x, fill=green!50, opacity=0.4, draw opacity=0, area legend, forget plot]   table [x expr=\thisrow{logu}-\thisrow{logl}, y=logloss]   {\tablesixheu} \closedcycle;
	\addplot [stack plots=x, stack dir=minus, forget plot, draw=none] table [x=logu, y=logloss] {\tablesixheu};
	
	\addplot [stack plots=x, fill=none, draw=none, forget plot]   table [x=logl, y=logloss]   {\tablesixxu} \closedcycle;
	\addplot [stack plots=x, fill=red!50, opacity=0.4, draw opacity=0, area legend, forget plot]   table [x expr=\thisrow{logu}-\thisrow{logl}, y=logloss]   {\tablesixxu} \closedcycle;
	\addplot [stack plots=x, stack dir=minus, forget plot, draw=none] table [x=logu, y=logloss] {\tablesixxu};
	
	\addplot [blue, only marks]  table [x=logm, y=logloss]   {\tablefive};
	\addplot [red, only marks]  table [x=logm, y=logloss]   {\tablefivexu};
	\addplot [green, only marks]  table [x=logm, y=logloss]   {\tablefiveheu};
	\addplot [black, only marks]  table [x=logm, y=logloss]   {\tablefivehen};

	\addplot [olive, solid, domain=-3:35]{0.5*(ln(0.5*0.9*0.2648/0.01))+0.5*x};
	
	\addplot [blue, only marks]  table [x=logm, y=logloss]   {\tablesix};
	\addplot [black, only marks]  table [x=logm, y=logloss]   {\tablesixhen};
	\addplot [green, only marks]  table [x=logm, y=logloss]   {\tablesixheu};
	\addplot [red, only marks]  table [x=logm, y=logloss]   {\tablesixxu};

	\legend{6-7 layer FC SN, 6-7 layer FC XU, 6-7 layer FC HU, 6-7 layer FC HN, Equation (\ref{eq:main})}
	
	\node[scale=0.75] at (axis cs: 0.5,15.5) {$\rho=0.9623$};
	
	\coordinate (c1) at (axis cs:-6,1.5);
	\coordinate (c2) at (axis cs:-6,3.5);
	\draw (c1) rectangle (axis cs:1,3.5);

	\end{axis}


\begin{axis}[name=ax2, at={($(ax1.south east)+(4.5cm,0)$)}, width=7cm, height=4cm, legend pos=outer north east,  legend style={fill=none, font=\footnotesize}, xlabel= {$\log(S_{\text{after}})$}, ylabel= { $\log(L)$}, every axis/.append style={label style={font=\small}}]

\addplot [stack plots=x, fill=none, draw=none, forget plot]   table [x=logl, y=logloss]   {\tablefivehen} \closedcycle;
\addplot [stack plots=x, fill=black!50, opacity=0.4, draw opacity=0, area legend, forget plot]   table [x expr=\thisrow{logu}-\thisrow{logl}, y=logloss]   {\tablefivehen} \closedcycle;
\addplot [stack plots=x, stack dir=minus, forget plot, draw=none] table [x=logu, y=logloss] {\tablefivehen};

\addplot [stack plots=x, fill=none, draw=none, forget plot]   table [x=logl, y=logloss]   {\tablefiveheu} \closedcycle;
\addplot [stack plots=x, fill=green!50, opacity=0.4, draw opacity=0, area legend, forget plot]   table [x expr=\thisrow{logu}-\thisrow{logl}, y=logloss]   {\tablefiveheu} \closedcycle;
\addplot [stack plots=x, stack dir=minus, forget plot, draw=none] table [x=logu, y=logloss] {\tablefiveheu};

\addplot [stack plots=x, fill=none, draw=none, forget plot]   table [x=logl, y=logloss]   {\tablefivexu} \closedcycle;
\addplot [stack plots=x, fill=red!50, opacity=0.4, draw opacity=0, area legend, forget plot]   table [x expr=\thisrow{logu}-\thisrow{logl}, y=logloss]   {\tablefivexu} \closedcycle;
\addplot [stack plots=x, stack dir=minus, forget plot, draw=none] table [x=logu, y=logloss] {\tablefivexu};

\addplot [stack plots=x, fill=none, draw=none, forget plot]   table [x=logl, y=logloss]   {\tablesixhen} \closedcycle;
\addplot [stack plots=x, fill=black!50, opacity=0.4, draw opacity=0, area legend, forget plot]   table [x expr=\thisrow{logu}-\thisrow{logl}, y=logloss]   {\tablesixhen} \closedcycle;
\addplot [stack plots=x, stack dir=minus, forget plot, draw=none] table [x=logu, y=logloss] {\tablesixhen};

\addplot [stack plots=x, fill=none, draw=none, forget plot]   table [x=logl, y=logloss]   {\tablesixheu} \closedcycle;
\addplot [stack plots=x, fill=green!50, opacity=0.4, draw opacity=0, area legend, forget plot]   table [x expr=\thisrow{logu}-\thisrow{logl}, y=logloss]   {\tablesixheu} \closedcycle;
\addplot [stack plots=x, stack dir=minus, forget plot, draw=none] table [x=logu, y=logloss] {\tablesixheu};

\addplot [stack plots=x, fill=none, draw=none, forget plot]   table [x=logl, y=logloss]   {\tablesixxu} \closedcycle;
\addplot [stack plots=x, fill=red!50, opacity=0.4, draw opacity=0, area legend, forget plot]   table [x expr=\thisrow{logu}-\thisrow{logl}, y=logloss]   {\tablesixxu} \closedcycle;
\addplot [stack plots=x, stack dir=minus, forget plot, draw=none] table [x=logu, y=logloss] {\tablesixxu};

\addplot [black, only marks]  table [x=logm, y=logloss]   {\tablefivehen};
\addplot [green, only marks]  table [x=logm, y=logloss]   {\tablefiveheu};
\addplot [red, only marks]  table [x=logm, y=logloss]   {\tablefivexu};

\addplot [black, only marks]  table [x=logm, y=logloss]   {\tablesixhen};
\addplot [green, only marks]  table [x=logm, y=logloss]   {\tablesixheu};
\addplot [red, only marks]  table [x=logm, y=logloss]   {\tablesixxu};


\node[scale=0.75] at (axis cs: -4,2.75) {$\rho=0.9284$};

\end{axis}
\draw [dashed] (c1) -- (ax2.south west);
\draw [dashed] (c2) -- (ax2.north west);

\end{tikzpicture}


\caption{Test loss $L$ versus sensitivity $S_\text{after}$ for networks trained on a subset of the CIFAR-10 training dataset where networks are initialized with different methods. On the right, we have a zoom in plot of the bottom left frame of the left figure.}
\label{fig:init}
\end{figure*}

	Another interesting observation is the effect of various parameter initialization techniques on the sensitivity and loss values, after the networks are trained (Fig.~\ref{fig:init}). %
	We consider four initialization techniques for network parameters in our experiments: (i) Standard Normal distribution (SN), (ii) Xavier \cite{glorot2010understanding} initialization method with uniform distribution (XU), (iii) He \cite{he2015delving} initialization method with uniform distribution (HU), and (iv) He initialization method with normal distribution (HN). As shown in Fig.~\ref{fig:init}, the relation between sensitivity $S_\text{after}$ and test loss~$L$ provides us with a new viewpoint on the success of the state-of-the-art initialization techniques; HN has the best generalization performance, alongside the lowest sensitivity value (the black points in Fig.~\ref{fig:init}).

	\subsection{Sensitivity of Untrained Networks as a Proxy for Generalization Loss}\label{sec:untrain}

	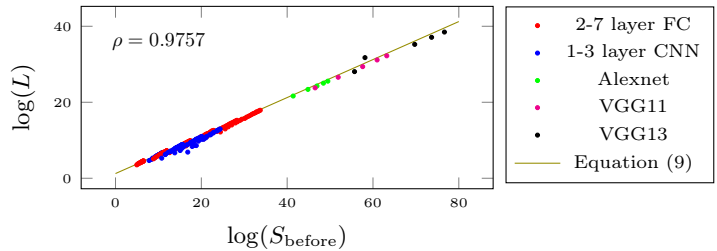
\begin{figure}[h]
	\caption{Test loss of trained models, $L$, versus sensitivity of untrained models, $S_\text{before}$, for networks whose parameters are initially drawn from the standard normal distribution. Note that the regularization techniques BN, dropout and max-pooling are removed from Alexnet, VGG11, and VGG13 configurations.}\label{fig:before}

		\begin{tikzpicture}[every mark/.append style={mark size=0.75pt}]
		
		\begin{axis}[width=7cm, height=4cm,  legend pos=outer north east, legend style={font=\scriptsize}, xlabel= {$\log(S_\text{before})$}, ylabel= { $\log(L)$}, cycle list name=color list]{

	\addplot  [red, only marks] table [x=logm, y=logloss]   {\tableone};

	\addplot  [blue, only marks] table [x=logm, y=logloss]   {\tableonec};

	\addplot [green, only marks]  table [x=logm, y=logloss]   {\tablealexnet};
	\addplot  [magenta, only marks] table [x=logm, y=logloss]   {\tablevggeleven};
	\addplot  [black, only marks] table [x=logm, y=logloss]   {\tablevggthirdteen};
	
	\addplot [olive, solid, domain=0:80]{0.5*(ln(0.5*0.9*0.2648/0.01))+0.5*x};

	\addplot  [red, only marks] table [x=logm, y=logloss]   {\tabletwo};
	\addplot  [red, only marks] table [x=logm, y=logloss]   {\tablethree};
	\addplot  [red, only marks] table [x=logm, y=logloss]   {\tablefour};
	\addplot  [red, only marks] table [x=logm, y=logloss]   {\tablefive};
	\addplot  [red, only marks] table [x=logm, y=logloss]   {\tablesix};
	
	\addplot  [blue, only marks] table [x=logm, y=logloss]   {\tabletwoc};
	\addplot  [blue, only marks] table [x=logm, y=logloss]   {\tablethreec};
	\addplot  [blue, only marks] table [x=logm, y=logloss]   {\tablefourc};
	
	\legend{2-7 layer FC, 1-3 layer CNN, Alexnet, VGG11, VGG13, Equation (\ref{eq:main})}
	\node[scale=0.75] at (axis cs: 10,36) {$\rho=0.9757$};

}
	\end{axis}
	
	\end{tikzpicture}

\end{figure}

	A similar trend is observed for neural networks that are not yet trained. In Fig.~\ref{fig:before}, the sensitivity $S_\text{before}$ is measured before the networks are trained, and the test loss $L$ is measured after the networks are trained. The parameters in the fully-connected and convolutional networks are initialized by sampling from the standard normal distribution, and no explicit regularization (dropout, BN, max-pooling) is used in the training process. These two conditions are necessary, because regularization techniques only affect the training process, hence $S_\text{before}$ is the same for networks with or without regularization layers, and the He and Xavier initialization techniques force the sensitivity to be the same regardless of the number of units in hidden layers. %
	Therefore, under these two conditions, the generalization performance of untrained networks with different architectures can be compared. 
	The strong link between the sensitivity of untrained networks~$S_\text{before}$ and the test loss $L$ observed in Fig.~\ref{fig:before} suggests that the generalization of neural networks can be compared before the networks are even trained, making sensitivity a computationally inexpensive architecture-selection method.

	\begin{figure}[h]
		\caption{Test loss $L_{\text{MSE}}$ versus sensitivity $S_\text{after}$ for a regression task with the MSE loss criterion. The networks are trained and evaluated on the Boston house price dataset. 
			Each point of the plot indicates a network with a different width and its coordinates are averaged over 10 runs.}\label{fig:reg2}

\begin{tikzpicture}[every mark/.append style={mark size=1pt}]

\begin{axis}[width=7cm, height=4cm, legend pos=outer north east, legend style={fill= none, font=\scriptsize}, xlabel= {$\log(S_\text{after})$}, ylabel= {$\log(L_{\text{MSE}})$ }]

\addplot [red, only marks]  table [x=logms, y=logloss]   {\tableonep};
\addplot [blue, only marks]  table [x=logms, y=logloss]   {\tabletwop};
\addplot [green, only marks]  table [x=logms, y=logloss]   {\tablethreep};
\addplot [cyan, only marks]  table [x=logms, y=logloss]   {\tablefourp};
\addplot [magenta, only marks]  table [x=logms, y=logloss]   {\tablefivep};
\addplot [black, only marks]  table [x=logms, y=logloss]   {\tablesixp};


\legend{3 layer FC, 4 layer FC, 5 layer FC, 6 layer FC, 7 layer FC, 8 layer FC, $L_\text{MSE}=S \frac{\sigma_{x}^2}{\sigma_{\varepsilon_x}^2}$}

\node[scale=0.75] at (axis cs: 35,8) {$\rho=0.7681$};

\end{axis}

\end{tikzpicture}

	\end{figure}
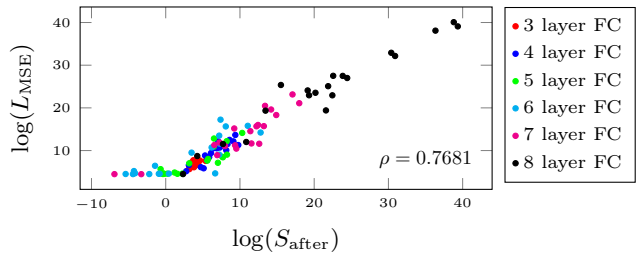

	\subsection{Regression Task and MSE Loss}\label{app:mse}
	In this section, we investigate the relation between sensitivity and generalization error for regression tasks with the mean square error criterion (MSE). 
	The loss function in this setting is defined in~(\ref{eq:msetestloss}) where $\theta^*$ is found by minimizing MSE on training dataset~$\mathcal{D}_t$ using the stochastic learning algorithm~$\mathcal{A}$. Note that the output is the last layer of the neural network (the softmax layer is not applied), and that the output layer has 1 unit, i.e.,~$K=1$, and that $y$ is a scalar. The sensitivity is defined as 
	$
		S=\E_{\theta^*}\left[ \mathrm{Var}_{x, \varepsilon_x}\left[F_{\theta^*}(x+\varepsilon_x) - F_{\theta^*}(x)\right]\right]
	$
	and the bias and variance terms are defined by~(\ref{eq:bias}) and (\ref{eq:var}), respectively.
	We consider the Boston housing dataset where the objective is to predict the price of a house given 14 features (including crime rate, distance to employment centers, etc.).
	Fig.~\ref{fig:reg2} shows sensitivity versus test loss among fully-connected neural networks with 3-8 layers and 100-500  hidden units per layer; the networks are trained on $70\%$ of the dataset and then evaluated on the remaining $30\%$. %
	The results are consistent with the relation between sensitivity $S$ and generalization error, which for the regression task is $L_\text{MSE}$. For a more detailed view, we observe that sensitivity is related to the variance in the bias-variance decomposition of the MSE loss (Fig.~\ref{fig:reg2_long}~(d) in the appendix), and the MSE loss is the sum of bias and variance terms (Fig.~\ref{fig:reg2_long}~(c) in the appendix).

\section{Discussion}\label{sec:dis}

\subsection{Discussion Regarding Bias}\label{app:bias}
In this section, we discuss the role of the number of training samples and of the stage of the training on the validity of the approximation made in (\ref{eq:main}) where we neglect $\varepsilon_{\textrm{bias}}$. In our experiments, we observe that when the number of training samples is low (see for instance Fig.~\ref{fig:bias_long}~(a) for ResNet18 and ResNet34 networks), the match between experiments and (\ref{eq:main}) is rather poor. We show (in Fig.~\ref{fig:bias_long}~(b) in the appendix) that this problem can be solved (at least in part) by training the networks with more samples: for instance, in Fig.~\ref{fig:bias_long}~(b) the yellow marks are ResNet18, the green marks are ResNet34, and the results show a relation between $\log S_{\text{after}}$ and $\log L$ that becomes linear as 
we add more training data samples in the training process. Therefore, the larger the number of training samples is, the better the approximation $\varepsilon_{\textrm{bias}} \approx train loss$ becomes, and $\varepsilon_{\textrm{variance}}$ becomes the more dominant term in the test loss. We also observe that, when computing sensitivity and loss at different stages of training, the bias term $\varepsilon_{\textrm{bias}}$ in the test loss cannot be neglected at initial stages of the training. As the training progresses, the experimental results get closer to~(\ref{eq:main}) (see Fig.~\ref{fig:bias_long}~(d) in the appendix).

\subsection{Final Remarks}
As discussed in Section~\ref{sec:bias_var}, the loss can be decomposed in three terms: $\varepsilon_{\textrm{variance}}$, $\varepsilon_{\textrm{bias}}$, and $\varepsilon_{\textrm{noise}}$. The proposed relation between sensitivity $S$ and variance $\varepsilon_{\textrm{variance}}$ is extended to a relation between $S$ and generalization loss $L$ when $\varepsilon_{\textrm{variance}}$ is the dominant term in the decomposition of the loss, which is often the case in deep learning settings. In the previous section, we discussed the possibility that the bias term $\varepsilon_{\textrm{bias}}$ might not be negligible compared to $\varepsilon_{\textrm{variance}}$, when the number of training samples is low and when the training loss is large. When the available data contains randomly labeled samples, then $\varepsilon_{\textrm{noise}}$ can no longer be neglected. As $S$ does not depend on the labels, the randomness in the labels, and therefore the generalization performance of the model, can no longer be entirely captured by sensitivity in this setting. 
Furthermore, the pixel-wise linear input perturbations considered in our experiments might not be realistic; ideally, we would like to perturb the input in the latent space of the generative model of the input image. Also, the relation between $S$ and $L$ requires the non-linearity to be positive homogeneous. The generalization properties of networks with sigmoid and tanh activation functions are left for future work.

The sensitivity $S$ changes with input-output re-scaling: For a homogeneous predictor, if the input data scale is multiplied by a factor $\alpha$, and the output is divided by the factor $\alpha$, then $L$ remains unchanged, whereas~$S$ gets divided by $\alpha^4$. However, as long as we compare networks subject to the same input data distribution, this re-scaling obviously does not happen. Moreover, $S$ can be affected by output re-scaling: If the output of a classifier is divided by a factor $\alpha$, then the classification accuracy remains the same, whereas $L$ and $S$ get divided by (approximately) $\alpha$ and $\alpha^2$, respectively. While the relation between $L$ and $S$ remains valid, there is a mismatch between accuracy and loss, which suggests that the networks are miscalibrated.
Applying network calibration methods such as \emph{temperature scaling}~\cite{guo2017calibration} can potentially increase the correlation between the cross-entropy loss $L$ and the classification error (i.e., {1 - accuracy}), as well as the correlation between the sensitivity $S$ and the classification error (see e.g., Table~\ref{tab:cal} in the appendix).

The relation between sensitivity and variance can be extended to any loss that admits a bias-variance decomposition. Therefore, if such a decomposition is found for the classification error (which might not be purely additive \cite{domingos2000unified}), which is still an active research topic, then the link between sensitivity and error would follow. We note that there is a difference between causality and correlation between a complexity measure and generalization, as discussed in \cite{jiang2019fantastic}. We study the correlation between sensitivity $S$ and generalization loss $L$, however, this does not imply that there is a causal relation between the two. %

\section{Conclusion}\label{sec:conclusion}

We find that the sensitivity metric is a strong indicator of overfitting. Given multiple networks having near-zero training losses to choose from with different hyper-parameters, the best architecture appears to be the one with the lowest sensitivity value. Sensitivity can also potentially be used as an architecture-selection method. 
One of the advantages of the sensitivity metric is that it can provide a loose prediction of the test loss without the use of any labeled data. This is especially important in applications where labeling data is expensive. 


\bibliographystyle{ieeetr}
\bibliography{main}

\onecolumn
\newpage
\appendix

	\section{Experimental Details}\label{app:exp}
	The CIFAR-10\footnote{\url{https://www.cs.toronto.edu/~kriz/cifar.html}} and the Boston house pricing\footnote{\url{https://www.cs.toronto.edu/~delve/data/boston/bostonDetail.html}} datasets are used for the image-classification and regression experiments presented in the main part of the paper. %
	The fully-connected neural networks have the same number of units in the hidden layers, varying from 100 to 500 with a step size of 20. For the convolutional neural networks the number of channels in convolutional layers vary from 5 to 25 with a step size of 5 (note that each time a channel is added in the convolutional layers, an extra 20 units is added to the fully-connected layers of the CNN accordingly). %
	As it is computationally expensive to reach zero training loss for the entire dataset, we choose a randomly sampled subset of the training set containing 1000 samples of the CIFAR-10 dataset. Zero training loss is necessary for a fair comparison between different networks since we would like to have the same value for $\varepsilon_{\textrm{bias}}$ and $\varepsilon_{\textrm{noise}}$ among them. %
	For the optimization algorithm, we choose the Adam optimizer with $lr=0.001$ and $betas=(0.9,0.999)$. We initialize the weights and biases with random values drawn from the distribution stated in each figure. The non-linear activation function is set to be the ReLU function throughout the experiments. 
	We stop the training when the training loss reaches below the threshold~$10^{-5}$ for 10 times. 
	In case this condition is not met, we stop the training after 2000 epochs (each epoch is iterations over the mini-batches with size 128 of the training set). The noise added to the input image is a random tensor with the same size as the input and is drawn from the Gaussian distribution with zero mean and $0.1$ standard deviation. The output noise is first averaged over all its $K$ entries (for CIFAR-10 the number of classes is  $K=10$), then we take its variance over inputs of the testing dataset and the input noise. All the reported experimental results are averaged over 10 runs.  Each experiment took few hours on one Nvidia Titan X Maxwell GPU. 
	
	We use the notations: 
	\begin{itemize}
		\item Conv(number of filters, kernel size, stride, padding)
		\item Maxpool(kernel size)
		\item Linear(number of units)
		\item Dropout(dropout rate)
	\end{itemize}
	for layers of a convolutional neural network where Conv and Linear layers also include the ReLU non-linearity except the very last linear layer. The configurations that are used are:
	\begin{itemize}
		\item The Alexnet \cite{krizhevsky2012imagenet}: Conv(h, 3, 2, 1) - Maxpool(2) - Conv(3*h, 3, 1, 1) - Maxpool(2) - Conv( 6*h, 3, 1, 1) - Conv(4*h, 3, 1, 1) - Conv(4*h, 3, 1, 1) - Maxpool(2) - Dense layer - Dropout(0.5) - Linear(4096) - Dropout(0.5) - Linear(4096) - Linear(K)\\ where $\text{h} \in [16, 32, 48, 64, 80]$
		\item VGG13 \cite{simonyan2014very} : 2 x Conv(64*s, 3, 1, 1) - Maxpool(2) - 2 x Conv(128*s, 3, 1, 1) - Maxpool(2) - 2 x Conv(256*s, 3, 1, 1) - Maxpool(2) - 2 x Conv(512*s, 3, 1, 1) - Maxpool(2) - 2 x Conv(512*s, 3, 1, 1) - Maxpool(2) - Avgpool(2) - Dense layer - Linear(K)\\ where $\text{s} \in [0.25, 0.5, 1, 1.5, 2]$ and all Conv layers have batch normalization
		\item Each block of a ResNet \cite{he2016deep} configuration: 2 x Conv(h, 3, 1, 1) + Conv(h, 1, 1, 1) which Conv layers include BN and ReLU and the result of the summation goes into a ReLU layer and h is the number of channels.
	\end{itemize}
	VGG16 is the same as VGG13 with the difference that it has three layers in the last three blocks. VGG11 configuration is the same as VGG13 except that in the first and second block it has one convolutional layer instead of 2. VGG19 is the same as VGG13 except that there is 4 conv layers instead of 2 in the last three blocks.
	ResNet18 has 2 blocks with h=64*s, 2 blocks with h=128*s, 2 blocks with h=256*s, and 2 blocks with h=512*s where $\text{s} \in [0.25, 0.5, 1, 1.5, 2]$. ResNet34 has 3 blocks with h=64*s, 4 blocks with h=128*s, 6 blocks with h=256*s, and 3 blocks with h=512*s where $\text{s} \in [0.25, 0.5, 1, 1.5, 2]$.

	\section{Computation of (\ref{eq:prop1}): The Relation between Variance and Sensitivity}\label{app:prop1}
	Computations of this section do not depend on the stage of the training, hence $\theta$ denotes the parameter vector at any stage of training.
	Let us recall the sensitivity metric (Equation (\ref{eq:sen})) definition
	\begin{equation*}
	S = \E_{\theta}\left[\mathrm{Var}_{x,\varepsilon_x}\left[\overline{\varepsilon_y}\right]\right] ,
	\end{equation*}
	where $\overline{\varepsilon_y} = 1/K \sum_{k=1}^{K} \varepsilon_y^k$, $\varepsilon_y^k$ is the $k$-th entry of output noise vector $\varepsilon_y$ given by
	\begin{equation*}
		\varepsilon_y^k = f_\theta^{k}(x+\varepsilon_x) - f_\theta^{k}(x) \approxeq \varepsilon_x \cdot \nabla_x^{\top} f_\theta^{k}(x) ,
	\end{equation*}
	where we apply a first order Taylor expansion of the output. For a one hidden layer fully-connected neural network with $D$ input units, $H$ hidden units, and $K$ output units, we have $\theta = \{w_1 \in \R^{D \times H}, w_2 \in \R^{H \times K}, {b_1 \in \R^{H}}, b_2\in \R^{K}\}$ where $w_l$ and $b_l$ are the weights and biases of layer $l$ ($l=1$ is the hidden layer and $l=2$ is the output layer), which are independently drawn from a zero-mean normal distribution: $w_1 \sim \mathcal{N}(0, \sigma^2_{w_1}I)$, $w_2 \sim \mathcal{N}(0, \sigma^2_{w_2}I)$, $b_1 \sim \mathcal{N}(0, \sigma^2_{b_1}I)$, and $b_2 \sim \mathcal{N}(0, \sigma^2_{b_2}I)$ (this assumption has been studied in  \cite{bellido1993backpropagation}). We have
	\begin{equation*}
	f^k_\theta(x) = \sum_{h=1}^{H} w_2^{hk} a(p^h) + b_2^k ,
	\end{equation*}
	where $w_l^{ij}$ is the weight connecting unit $i$ in layer $l$ to unit $j$ in layer $l+1$, $b_l^h$ is the bias term added to unit $h$ in layer $l+1$, $p^h$ is the output of the linear transformation in the hidden unit $h$, i.e., 
	\begin{equation*}
		p^h = \sum_{d=1}^{D} w_1^{dh} x^d + b_1^h ,
	\end{equation*}
	and the non-linear activation function $a(\cdot)$ is a positive homogeneous function of degree $1$; i.e.,
	\begin{equation}\label{eq:nonl}
	a(x) = 
	\begin{cases}
	\alpha x & x>0, \\
	\beta x & \text{otherwise},
	\end{cases}
	\end{equation}
	where $\alpha$ and $\beta$ are non-negative hyper-parameters. ReLU follows~(\ref{eq:nonl}) with $\alpha=1$ and $\beta=0$. By applying the chain rule we obtain
	\begin{align*}
		\varepsilon_y^k \approx \sum_{d=1}^{D} \varepsilon_x^{d} \frac{\partial f^k_\theta(x)}{\partial x^d} 
		=  \sum_{d=1}^{D} \varepsilon_x^{d} \sum_{h=1}^{H} w_2^{hk} w_1^{dh} \frac{\partial a(p^h)}{\partial p^h} .
	\end{align*}
	Therefore, we have
	\begin{equation*}
	\overline{\varepsilon_y} = \frac{1}{K} \sum_{k=1}^{K} \sum_{h=1}^{H} \sum_{d=1}^{D} \varepsilon_x^d w_2^{hk} w_1^{dh} \frac{\partial a(p^h)}{\partial p^h} .
	\end{equation*}
	The network parameters are assumed to be independent from each other, and it is assumed that $x \independent \theta$, and~$\varepsilon_x \independent \{\theta, x\}$. Moreover, the entries of the input vector $x$ are independent from each other with the same second moment, i.e.,~$\sigma^2_x=\E[(x^d)^2]$ for~$1 \leq d \leq D$. Consider the input noise $\varepsilon_x$ to be a vector of zero mean random variables, hence~$S = \E_{\theta, x, \varepsilon_x} [\left(\overline{\varepsilon_y}\right)^2]$.
	Then the sensitivity becomes
	\begin{align}\label{eq:sencal}
	S &= \frac{1}{K^2} \sum_{k=1}^{K} \sum_{h=1}^{H} \sum_{d=1}^{D} \E_{\varepsilon_x}[(\varepsilon_x^d)^2] \E_{\theta, x} \left[ (w_2^{hk})^2 (w_1^{dh})^2 \left( \frac{\partial a(p^h)}{\partial p^h} \right)^2 \right] \nonumber \\
	&= \frac{1}{K^2} \sum_{k=1}^{K} \sum_{h=1}^{H} \sum_{d=1}^{D} \sigma^2_{\varepsilon_x} \sigma^2_{w_2} \sigma^2_{w_1} \frac{\alpha^2+\beta^2}{2} = \frac{HD}{K} \sigma^2_{\varepsilon_x} \sigma^2_{w_2} \sigma^2_{w_1} \frac{\alpha^2+\beta^2}{2} ,
	\end{align}
	where the second equation follows by computing the expectation for zero-mean normal parameters. Let
	\begin{equation}\label{eq:defvar}
	var = \E_x \left[ \textrm{Var}_\theta [out] \right] ,
	\end{equation}
	where $out = 1/K \sum_{k=1}^{K} f_\theta^k(x)$. Because of the homogeneity of the non-linearity $a(\cdot)$, we have $a(p^h) = p^h \cdot \frac{\partial a(p^h)}{\partial p^h}$. Hence
	\begin{equation*}
		out = \frac{1}{K} \sum_{k=1}^{K} \left[\sum_{h=1}^{H} w_2^{hk}  \left( \sum_{d=1}^{D} w_1^{dh} x^d + b_1^h \right) \frac{\partial a(p^h)}{\partial p^h} + b_2^k\right] .
	\end{equation*}
	Because the parameters are zero-mean, $var = \E_{\theta, x} \left[out^2\right]$ and we have
	\begin{align*}
	var &= \frac{1}{K^2} \sum_{k=1}^{K} \sum_{h=1}^{H} \sum_{d=1}^{D}  \E \left[ (x^d)^2 (w_2^{hk})^2 (w_1^{dh})^2 \left( \frac{\partial a(p^h)}{\partial p^h} \right)^2 \right] \\
	&+ \frac{1}{K^2} \sum_{k=1}^{K} \sum_{h=1}^{H} \E\left[(w_2^{hk})^2\right] \E\left[ (b_1^h)^2 \left( \frac{\partial a(p^h)}{\partial p^h} \right)^2 \right] + \frac{1}{K^2} \sum_{k=1}^{K} \E\left[(b_2^k)^2\right] \\
	&= \frac{1}{K^2} \sum_{k=1}^{K} \sum_{h=1}^{H} \sum_{d=1}^{D} \sigma^2_x \sigma^2_{w_2} \sigma^2_{w_1} \frac{\alpha^2+\beta^2}{2} \\
	&+ \frac{1}{K^2} \sum_{k=1}^{K} \sum_{h=1}^{H} \sigma^2_{w_2} \sigma^2_{b_1}  \frac{\alpha^2+\beta^2}{2} +  \frac{1}{K^2} \sum_{k=1}^{K} \sigma^2_{b_2} \\
	&= \frac{HD}{K} \sigma^2_x \sigma^2_{w_2} \sigma^2_{w_1} \frac{\alpha^2+\beta^2}{2} + \frac{H}{K} \sigma^{2}_{{w_2}} \sigma^{2}_{{b_1}} \frac{\alpha^2+\beta^2}{2} + \frac{1}{K} \sigma^{2}_{{b_2}} ,
	\end{align*}
	which follows by taking the expectations over the parameters with zero-mean normal distributions. Therefore, we obtain
	\begin{equation*}
	var = S \cdot \frac{\sigma^{2}_{{x}}}{\sigma^{2}_{{\varepsilon_x}}} + \frac{H}{K} \sigma^{2}_{{w_2}} \sigma^{2}_{{b_1}} \frac{\alpha^2+\beta^2}{2} + \frac{\sigma^{2}_{{b_2}}}{K} , 
	\end{equation*}
	where $\sigma^{2}$ denotes the second moment of a random variable. 
	Following the same computations for a neural network with $M$ hidden layers, we have	
	\begin{align}\label{eq:varsen}
	var &= S \cdot \frac{\sigma^{2}_{{x}}}{\sigma^{2}_{{\varepsilon_x}}} 
	+ \frac{1}{K} \sum_{l=1}^{M} \sigma^{2}_{{b_l}} \prod_{i=l+1}^{M} \frac{\alpha^2+\beta^2}{2} \sigma^{2}_{{w_{i}}} H_{i} ,
	\end{align}
	where $K$ is the number of units of the output layer $M+1$. We refer to the second term in the right-hand side of~(\ref{eq:varsen}) as $\Sigma$. Its value is a very rough approximation given the numerous assumptions made above, but in practice it can often be neglected because $\sigma_{b_l}^2$ is very small or zero (the ResNet configurations do not have biases) in most of our experiments. So far, an approximate relation between sensitivity and variance before the softmax function was established. Next, we find a relation between sensitivity before the softmax $S$ and variance after the softmax layer~$\varepsilon_{\textrm{variance}}$.\\%
	The first order Taylor expansion for an arbitrary function at the average of its input is $${g(x) \approx g(\E[x])  + g^{\prime}(\E[x]) (x-\E[x])}.$$ Taking the variance of $g(x)$, we have
	\begin{equation*}
	\textrm{Var}(g(x)) \approx  \left( g^{\prime}(\E[x]) \right)^2 \textrm{Var}(x).
	\end{equation*}
	Here the function $g(\cdot)$ is the softmax function with input vector $f_\theta(x)$ and output indices
	\begin{equation*}
		F^k_\theta(x) = \frac{\exp(f^k_\theta(x))}{\sum_{i=1}^{K} \exp(f_\theta^{i}(x))} ,
	\end{equation*}
	for $1 \leq k \leq K$.
	The input of the softmax function is a $K$-dimensional vector, so the variance of the output includes the vector-matrix multiplication of the covariance matrix of the input and the gradient vector. We assume that the outputs of the last layer are independent from each other ($f_\theta^{i} \independent f_\theta^{j}$ for $1 \leq i,j \leq K, i \neq j$), so the covariance matrix is a diagonal matrix. Because the parameters are considered to be zero-mean, the input of the softmax has zero mean, $\E[f^k_\theta(x)]=0$ for $1 \leq k \leq K$, then
	\begin{align*}
	\textrm{Var}(F^k_\theta(x)) & \approx \sum_{i=1}^{K} \left(\left.\frac{\partial F^k_\theta(x)}{\partial f^i_\theta(x)}\right\vert_{\E[f^k_\theta(x)]=0}\right)^2 \textrm{Var}(f^{i}_\theta(x))\\
	&\approx \left(\frac{1}{K}\cdot\left(1-\frac{1}{K}\right)\right)^2 \textrm{Var}(f^k_\theta(x)) + \left(-\frac{1}{K^2}\right)^2 \sum\limits_{\substack{i=1 \\ i\neq k}}^K \textrm{Var}(f^i_\theta(x)) ,
	\end{align*}
	as $\text{softmax}(0)=1/K$.
	Therefore,
	\begin{align*}
	\varepsilon_{\textrm{variance}} = \sum_{k=1}^{K} \E_x \left[\textrm{Var}(F^k_\theta(x))\right] \approx  K \left(\frac{(K-1)^2}{K^4} + \frac{K-1}{K^4}\right)  K \cdot var 
	= \left(\frac{K-1}{K}\right) \left(S \cdot \frac{\sigma^{2}_{{x}}}{\sigma^{2}_{{\varepsilon_x}}}  + \Sigma \right)  ,
	\end{align*}
	which completes the computations.
	\section{The Relation between the Cross Entropy Loss and the Mean Square Error}\label{app:prop2}
	We rewrite the cross-entropy loss (Equation (\ref{eq:cetestloss})) as
	\begin{equation*}
	L = \E_{\theta^*}[L_{\theta^*}] = \E_{x,c,\theta^*}\left[-\log(F_{\theta^*}^c)\right] ,
	\end{equation*}
	where $1 \leq c \leq K $ is the index of the true class for the input $x$, i.e., $y^c=1$ and $y^k=0$ for $k \neq c$ . For simplicity we use the notation $F_{\theta^*}^c$ instead of $F_{\theta^*}^c(x)$ in this section. For the MSE loss we have
	\begin{equation*}
	L_{\textrm{MSE}} = \E_{x,y,\theta^*} \left[\sum_{k=1}^{K} \left(F_{\theta^*}^k-y^k\right)^2\right].
	\end{equation*}
	Because $\sum\limits_{\substack{k=1}}^{K} F_{\theta^*}^k = F_{\theta^*}^c + \sum\limits_{\substack{j=1 \\ j\neq c}}^K F_{\theta^*}^j = 1$ the summation inside the above expectation, can be rewritten by replacing~$y^k$ by their $0-1$ values
	\begin{align*}
	\sum_{k=1}^{K} \left(F_{\theta^*}^k-y^k\right)^2 &=  \left(1-F_{\theta^*}^c\right)^2 + \sum\limits_{\substack{j=1 \\ j\neq c}}^K   \left(F_{\theta^*}^j\right)^2    
	= \left(1-F_{\theta^*}^c\right)^2  +  \left(1-F_{\theta^*}^c\right)^2 - \sum\limits_{\substack{i=1 \\ i\neq c}}^K\sum\limits_{\substack{j=1 \\ j\neq i,c }}^K F_{\theta^*}^{i} F_{\theta^*}^j \\
	&= 2\left(1-F_{\theta^*}^c\right)^2 - \sum\limits_{\substack{i=1 \\ i\neq c}}^{K}\sum\limits_{\substack{j=1 \\ j\neq i,c }}^K F_{\theta^*}^{i} F_{\theta^*}^j .
	\end{align*}
	Since $0 \leq F_{\theta^*}^j \leq (1-F_{\theta^*}^c)$ for $1 \leq j \leq K, j \neq c$ and $\sum\limits_{\substack{j=1 \\ j\neq c}}^K F_{\theta^*}^j = 1-F_{\theta^*}^c$, the above equation is bounded by
	\begin{equation}\label{eq:ineq}
		\left(\frac{K}{K-1}\right)\left(1- F_{\theta^*}^c\right)^2 \leq \sum_{k=1}^{K} \left(F_{\theta^*}^k-y^k\right)^2 \leq 2 \left(1-F_{\theta^*}^c\right)^2 .
	\end{equation}
	The lower bound in the above inequality occurs when $F_{\theta^*}^j = (1- F_{\theta^*}^c)/(K-1)$ for $1 \leq j \leq K, j \neq c$ and the upper bound above occurs when all the remaining probability (i.e., $1-F_{\theta^*}^c$) is given to one class besides the true class $c$, and the rest of the classes are assigned with zero probability. %
	The inequality in Equation (\ref{eq:ineq}) can be rewritten in the following inequality
	\begin{equation}\label{eq:here}
	\sqrt{\frac{1}{2}  \sum_{k=1}^{K} \left(F_{\theta^*}^k-y^k\right)^2} \leq 1 - F_{\theta^*}^c \leq \sqrt{\frac{K-1}{K}  \sum_{k=1}^{K} \left(F_{\theta^*}^k-y^k\right)^2} .
	\end{equation}
	Intuitively, the upper bound above is preferable in practice, because we would like the network to be less confident in assigning probabilities to wrong classes. If we take expectations in~(\ref{eq:here}) and apply Jensen's inequality, the upper bound is upper bounded by $\sqrt{\frac{K-1}{K}} \sqrt{L_\text{MSE}}$. However, in our experiments, we often observe that the network assigns the probability $1 - F_{\theta^*}^c$ to a wrong class and zero to the remaining classes, i.e., the network is over-confident. Therefore, if we consider this scenario, we then approximate $1 - F_{\theta^*}^c$ with the lower bound above.
	Hence, by approximating the expectation of a squared root with the squared root of expectation, and by applying a first order Taylor expansion for the logarithm, i.e., 
	$
	{- \log (F_{\theta^*}^c) \approx 1 - F_{\theta^*}^c ,}
	$ we have\footnote{Note that if instead we would have considered the scenario that the network is not over-confident, then by approximating $1-F_{\theta^*}^c$ with the upper bound of inequality~(\ref{eq:here}), we would have had ${L \approx \sqrt{\frac{(K-1) L_\text{MSE}}{K}}}$, which differs from~(\ref{eq:here2}) by only a constant scaling factor.}
	\begin{equation}\label{eq:here2}
	L \approx \sqrt{\frac{L_\text{MSE}}{2}} .
	\end{equation}

	\section{Computation of (\ref{eq:senfc})}\label{app:prop3}
	Consider a feedforward FC with ReLU activation function ($\alpha=1, \beta=0$) where i.i.d. zero mean random noise~$\varepsilon_x$ with variance~$\sigma_{\varepsilon_x}^2$ is added to the input. Then, assuming the output noise entries are independent from each other, we have
	\begin{equation*}
	S = \frac{1}{K^2}\sum_{k=1}^{K}\textrm{Var}\left[\varepsilon_y^k\right] = \frac{1}{K^2}\sum_{k=1}^{K}\E\left[(\varepsilon_y^k)^2\right] .
	\end{equation*}
	If we have $M$ hidden layers with $H_l, 1 \leq l \leq M$ units per layer, assuming the parameters are i.i.d. and independent from the input noise $\varepsilon_x$, and are drawn from the standard normal distribution, following the same computations as in~(\ref{eq:sencal}) for a network with $M$ hidden layers, $D$ input units and $K$ output units, 
	\begin{equation*}
		S = \frac{1}{K^2} \sum_{k=1}^{K} D \sigma^2_{\varepsilon_x} \prod_{l=1}^{M} \frac{H_l}{2} .
	\end{equation*}
	If all the hidden layers have the same number of units, $H_1 = H_{2} = \cdots = H_{M} = H$, then,
	\begin{equation*}
	S =  \frac{D}{K}\left(\frac{H}{2} \right)^M \sigma^2_{\varepsilon_x} .
	\end{equation*}

	\section{CIFAR-10 Experiments}\label{app:ciar10}
	Fig.~\ref{fig:effect_of_stuff} shows the effect of different initialization techniques, and of adding dropout and batch normalization layers to fully-connected and convolutional neural networks trained on 1000 samples of the CIFAR-10 training dataset, and evaluated on the entire CIFAR-10 testing dataset. We observe again the strong relation between sensitivity $S_\text{after}$ and generalization error $L$ and the effect of these techniques on both $S_\text{after}$ and $L$. In Fig.~\ref{fig:varsen}, we present the empirical results on the relation between $var$ defined in~(\ref{eq:defvar}) and $S$ defined in~(\ref{eq:sen}). We experiment for 5 cases, where we change the second moment of the input $\sigma_x^2$ and the input noise $\sigma_{\varepsilon_x}^2$. In Figs.~\ref{fig:varsen}~(a) and (b), the original CIFAR-10 images are considered and in Figs.~\ref{fig:varsen}~(c), (d) and~(e), we normalize the inputs accordingly to change $\sigma_{x}^2$. In all the figures, the empirical relation between $var$ and $S$ shows a good match with~(\ref{eq:varsen}) where $\Sigma$ is neglected.

	\begin{figure}[h]
		
		\subfloat[Effect of initialization]{
			\begin{tikzpicture}[every mark/.append style={mark size=1pt}]
			
			\begin{axis}[width=5.7cm, height=4.7cm, legend pos=outer north east, legend style={font=\footnotesize}, xlabel= {$\log(S_{\text{after}})$}, ylabel= { $\log(L)$}]{
			
			\addplot [teal, only marks]  table [x=logm, y=logloss]   {\tablefive};
			\addplot [magenta, only marks]  table [x=logm, y=logloss]   {\tablefivex};
			
			\addplot [olive, solid, domain=-4:25]{0.5*(ln(0.5*0.9*0.2648/0.01))+0.5*x};
			
			\addplot [teal, only marks]  table [x=logm, y=logloss]   {\tableconvtwo};
			\addplot [magenta, only marks]  table [x=logm, y=logloss]   {\tableconvtwox};
			\addplot [teal, only marks]  table [x=logm, y=logloss]   {\tableconvthree};
			\addplot [magenta, only marks]  table [x=logm, y=logloss]   {\tableconvthreex};
			\addplot [teal, only marks]  table [x=logm, y=logloss]   {\tableconvfour};
			\addplot [magenta, only marks]  table [x=logm, y=logloss]   {\tableconvfourx};

			\legend{networks SN, networks XU, Equation (\ref{eq:main})}}
		
			\node[scale=0.75] at (axis cs: 1,11) {$\rho=0.8845$};
			
			\end{axis}
			
			\end{tikzpicture}
		
		}
	\subfloat[Effect of adding dropout]{
		\begin{tikzpicture}[every mark/.append style={mark size=1pt}]
		
		\begin{axis}[width=5.7cm, height=4.7cm, legend pos=outer north east, legend style={font=\footnotesize}, xlabel= {$\log(S_\text{after})$}, ylabel= {$\log(L)$}]{
			
			\addplot [teal, only marks]  table [x=logm, y=logloss]   {\tablethree};
			\addplot [magenta, only marks]  table [x=logm, y=logloss]   {\tablethreed};
			
			\addplot [olive, solid, domain=-4:34]{0.5*(ln(0.5*0.9*0.2648/0.01))+0.5*x};
			
			\addplot [teal, only marks]  table [x=logm, y=logloss]   {\tablefive};
			\addplot [magenta, only marks]  table [x=logm, y=logloss]   {\tablefived};
			\addplot [teal, only marks]  table [x=logm, y=logloss]   {\tableseven};
			\addplot [magenta, only marks]  table [x=logm, y=logloss]   {\tablesevend};
			\addplot [teal, only marks]  table [x=logm, y=logloss]   {\tableconvone};
			\addplot [magenta, only marks]  table [x=logm, y=logloss]   {\tableconvoned};
			\addplot [teal, only marks]  table [x=logm, y=logloss]   {\tableconvtwo};
			\addplot [magenta, only marks]  table [x=logm, y=logloss]   {\tableconvtwod};
			\addplot [teal, only marks]  table [x=logm, y=logloss]   {\tableconvthree};
			\addplot [magenta, only marks]  table [x=logm, y=logloss]   {\tableconvthreed};
			\addplot [teal, only marks]  table [x=logm, y=logloss]   {\tableconvfour};
			\addplot [magenta, only marks]  table [x=logm, y=logloss]   {\tableconvfourd};

		\legend{networks, networks+dropout, Equation (\ref{eq:main})}}
	
		\node[scale=0.75] at (axis cs: 2,12) {$\rho=0.9131$};
		
		\end{axis}
		
		\end{tikzpicture}
	}\\
\subfloat[Effect of adding batch normalization]{
	\begin{tikzpicture}[every mark/.append style={mark size=1pt}]
	
	\begin{axis}[width=5.7cm, height=4.7cm, legend pos=outer north east, legend style={font=\footnotesize}, xlabel= {$\log(S_\text{after})$}, ylabel= {$\log(L)$}]{
		
		\addplot [teal, only marks]  table [x=logm, y=logloss]   {\tablethree};
		\addplot [magenta, only marks]  table [x=logm, y=logloss]   {\tablethreebn};
		
		\addplot [olive, solid, domain=-4:34]{0.5*(ln(0.5*0.9*0.2648/0.01))+0.5*x};
		
		\addplot [teal, only marks]  table [x=logm, y=logloss]   {\tablefive};
		\addplot [magenta, only marks]  table [x=logm, y=logloss]   {\tablefivebn};
		\addplot [teal, only marks]  table [x=logm, y=logloss]   {\tableseven};
		\addplot [magenta, only marks]  table [x=logm, y=logloss]   {\tablesevenbn};
		\addplot [teal, only marks]  table [x=logm, y=logloss]   {\tableconvone};
		\addplot [magenta, only marks]  table [x=logm, y=logloss]   {\tableconvonebn};
		\addplot [teal, only marks]  table [x=logm, y=logloss]   {\tableconvtwo};
		\addplot [magenta, only marks]  table [x=logm, y=logloss]   {\tableconvtwobn};
		\addplot [teal, only marks]  table [x=logm, y=logloss]   {\tableconvthree};
		\addplot [magenta, only marks]  table [x=logm, y=logloss]   {\tableconvthreebn};
		\addplot [teal, only marks]  table [x=logm, y=logloss]   {\tableconvfour};
		\addplot [magenta, only marks]  table [x=logm, y=logloss]   {\tableconvfourbn};

		\legend{networks, networks+BN, Equation (\ref{eq:main})}}
	
	\node[scale=0.75] at (axis cs: 3,13) {$\rho=0.9642$};
	
	\end{axis}
	
	\end{tikzpicture}
}
	\caption{Test loss $L$ versus sensitivity $S_\text{after}$ for networks trained on 1000 samples of the CIFAR-10 training dataset presenting the effect of initialization, dropout and batch normalization. Each point represents a different architecture and its coordinates are averaged over 10 runs. \textbf{(a)} The networks are 5 layer FC, 2-4 layer CNN where the parameters are initially drawn from either Xavier uniform distribution (XU) or standard normal distribution (SN). \textbf{(b)} The networks are 3, 5, 7 layer FC and 1-4 layer CNN. The top right most pink point is the same network architecture as the top right most teal blue point when dropout is added to the configuration. Hence, for all network architectures we observe a shift of the numerical points towards bottom left of the figure when dropout is applied. \textbf{(c)} The networks are 3, 5, 7 layer FC and 1-4 layer CNN. In (b) and (c) the networks parameters are initially drawn from the standard normal distribution.}
	\label{fig:effect_of_stuff}
\end{figure}
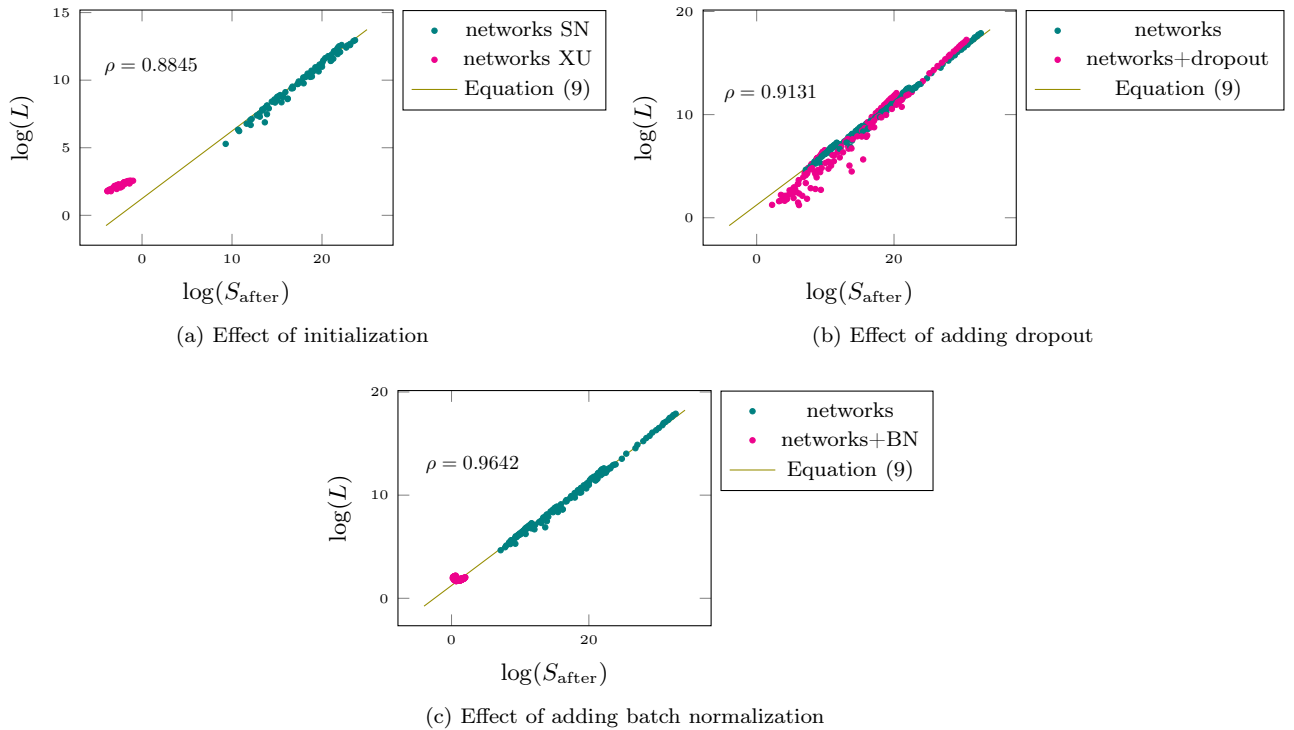

	\begin{figure}[h]
		
		\subfloat[$\sigma_x^2=0.2648$ and $\sigma_{\varepsilon_x}^2=0.01$]{
			\begin{tikzpicture}[every mark/.append style={mark size=0.9pt}]
			
			\begin{axis}[width=6.5cm, height=5.5cm, legend pos=outer north east, legend columns=2, legend style={font=\scriptsize}, xlabel= {$\log(S_{\text{after}})$}, ylabel= { $\log(var)$}]{
				
				\addplot [red, only marks]  table [x=logs, y=logvar]   {\tablethreeone};
				\addplot [green, only marks]  table [x=logs, y=logvar]   {\tablethreexone};
				\addplot [blue, only marks]  table [x=logs, y=logvar]   {\tablethreehnone};
				\addplot [magenta, only marks]  table [x=logs, y=logvar]   {\tablethreebnone};
				\addplot [yellow, only marks]  table [x=logs, y=logvar]   {\tablethreedone};
				\addplot [black, only marks]  table [x=logs, y=logvar]   {\tablefourone};
				\addplot [brown, only marks]  table [x=logs, y=logvar]   {\tablefourxone};
				\addplot [lime, only marks]  table [x=logs, y=logvar]   {\tablefourhnone};
				\addplot [orange, only marks]  table [x=logs, y=logvar]   {\tablefourbnone};
				\addplot [pink, only marks, mark=o]  table [x=logs, y=logvar]   {\tableresnetehnone};
				\addplot [olive, only marks, mark=o]  table [x=logs, y=logvar]   {\tableresnetthnone};
				\addplot [gray, only marks, mark=o]  table [x=logs, y=logvar]   {\tableresnetfhnone};
				\addplot [cyan, only marks, mark=triangle*]  table [x=logs, y=logvar]   {\tablevggehnone};
				\addplot [purple, only marks,mark=triangle*]  table [x=logs, y=logvar]   {\tablevggthnone};
				\addplot [violet, only marks, mark=triangle*]  table [x=logs, y=logvar]   {\tablevggshnone};
				\addplot [teal, only marks, mark=triangle*]  table [x=logs, y=logvar]   {\tablevggnhnone};
				\addplot [olive, solid, domain=-6:24]{(ln(0.2648/0.01))+x};

				\legend{3 layer FC SN, 3 layer FC XU, 3 layer FC HN, 3 layer FC SN BN, 3 layer FC SN dropout, 4 layer FC SN, 4 layer FC XU, 4 layer FC HN, 4 layer FC SN BN, ResNet18 HN, ResNet34 HN, ResNet50 HN, VGG11 HN, VGG13 HN, VGG16 HN, VGG19 HN, Equation (\ref{eq:varsen})}}
			
			\node[scale=0.75] at (axis cs: 0,16) {$\rho=0.9670$};
			
			\end{axis}
			
			\end{tikzpicture}
			
		}\\
	\subfloat[$\sigma_x^2=0.2648$ and $\sigma_{\varepsilon_x}^2=0.04$]{
		\begin{tikzpicture}[every mark/.append style={mark size=0.9pt}]
		
		\begin{axis}[width=6.6cm, height=5.5cm, legend columns = 2, legend style={font=\scriptsize, at={(1.12,1.43)}}, xlabel= {$\log(S_{\text{after}})$}, ylabel= { $\log(var)$}]{
			
			\addplot [red, only marks]  table [x=logs, y=logvar]   {\tablethreetwo};
			\addplot [green, only marks]  table [x=logs, y=logvar]   {\tablethreextwo};
			\addplot [blue, only marks]  table [x=logs, y=logvar]   {\tablethreehntwo};
			\addplot [magenta, only marks]  table [x=logs, y=logvar]   {\tablethreebntwo};
			\addplot [yellow, only marks]  table [x=logs, y=logvar]   {\tablethreedtwo};
			\addplot [cyan, only marks, mark=triangle*]  table [x=logs, y=logvar]   {\tablevggehntwo};
			\addplot [olive, solid, domain=-4:24]{(ln(0.2648/0.04))+x};
			
			\addplot [red, only marks]  table [x=logs, y=logvar]   {\tablefourtwo};
			\addplot [green, only marks]  table [x=logs, y=logvar]   {\tablefourxtwo};
			\addplot [blue, only marks]  table [x=logs, y=logvar]   {\tablefourhntwo};
			\addplot [magenta, only marks]  table [x=logs, y=logvar]   {\tablefourbntwo};

			\legend{3-4 layer FC SN, 3-4 layer FC XU, 3-4 layer FC HN, 3-4 layer FC SN BN, 3 layer FC SN dropout, VGG11 HN, Equation (\ref{eq:varsen})}}
		
		\node[scale=0.75] at (axis cs: 0,16) {$\rho=0.8910$};
		
		\end{axis}
		
		\end{tikzpicture}
		
	} \qquad
\subfloat[$\sigma_x^2=1$ and $\sigma_{\varepsilon_x}^2=0.01$]{
	\begin{tikzpicture}[every mark/.append style={mark size=0.9pt}]
	
	\begin{axis}[width=6.6cm, height=5.5cm, legend columns = 2, legend style={font=\scriptsize, at={(1.12,1.43)}}, xlabel= {$\log(S_{\text{after}})$}, ylabel= { $\log(var)$}]{
		
		\addplot [red, only marks]  table [x=logs, y=logvar]   {\tablethreethree};
		\addplot [green, only marks]  table [x=logs, y=logvar]   {\tablethreexthree};
		\addplot [blue, only marks]  table [x=logs, y=logvar]   {\tablethreehnthree};
		\addplot [magenta, only marks]  table [x=logs, y=logvar]   {\tablethreebnthree};
		\addplot [yellow, only marks]  table [x=logs, y=logvar]   {\tablethreedthree};
		\addplot [cyan, only marks, mark=triangle*]  table [x=logs, y=logvar]   {\tablevggehnthree};
		
		\addplot [olive, solid, domain=-8:24]{(ln(1/0.01))+x};
		
		\addplot [red, only marks]  table [x=logs, y=logvar]   {\tablefourthree};
		\addplot [green, only marks]  table [x=logs, y=logvar]   {\tablefourxthree};
		\addplot [blue, only marks]  table [x=logs, y=logvar]   {\tablefourhnthree};
		\addplot [magenta, only marks]  table [x=logs, y=logvar]   {\tablefourbnthree};

		\legend{3-4 layer FC SN, 3-4 layer FC XU, 3-4 layer FC HN, 3-4 layer FC SN BN, 3 layer FC SN dropout, VGG11 HN, Equation (\ref{eq:varsen})}}
	
	\node[scale=0.75] at (axis cs: 0,15) {$\rho=0.9501$};
	
	\end{axis}
	
	\end{tikzpicture}
	
}\\
\subfloat[$\sigma_x^2=1$ and $\sigma_{\varepsilon_x}^2=0.04$]{
	\begin{tikzpicture}[every mark/.append style={mark size=0.9pt}]
	
	\begin{axis}[width=6.6cm, height=5.5cm, legend columns = 2, legend style={font=\scriptsize, at={(1.12,1.43)}}, xlabel= {$\log(S_{\text{after}})$}, ylabel= { $\log(var)$}]{
		
		\addplot [red, only marks]  table [x=logs, y=logvar]   {\tablethreefour};
		\addplot [green, only marks]  table [x=logs, y=logvar]   {\tablethreexfour};
		\addplot [blue, only marks]  table [x=logs, y=logvar]   {\tablethreehnfour};
		\addplot [magenta, only marks]  table [x=logs, y=logvar]   {\tablethreebnfour};
		\addplot [yellow, only marks]  table [x=logs, y=logvar]   {\tablethreedfour};
		\addplot [cyan, only marks, mark=triangle*]  table [x=logs, y=logvar]   {\tablevggehnfour};
		\addplot [olive, solid, domain=-8:24]{(ln(1/0.04))+x};
		
		\addplot [red, only marks]  table [x=logs, y=logvar]   {\tablefourfour};
		\addplot [green, only marks]  table [x=logs, y=logvar]   {\tablefourxfour};
		\addplot [blue, only marks]  table [x=logs, y=logvar]   {\tablefourhnfour};
		\addplot [magenta, only marks]  table [x=logs, y=logvar]   {\tablefourbnfour};

		\legend{3-4 layer FC SN, 3-4 layer FC XU, 3-4 layer FC HN, 3-4 layer FC SN BN, 3 layer FC SN dropout, VGG11 HN, Equation (\ref{eq:varsen})}}
	
	\node[scale=0.75] at (axis cs: 0,16) {$\rho=0.9742$};
	
	\end{axis}
	
	\end{tikzpicture}
	
} \qquad
\subfloat[$\sigma_x^2=1.25$ and $\sigma_{\varepsilon_x}^2=0.01$]{
	\begin{tikzpicture}[every mark/.append style={mark size=0.9pt}]
	
	\begin{axis}[width=6.6cm, height=5.5cm, legend columns = 2, legend style={font=\scriptsize, at={(1.12,1.37)}}, xlabel= {$\log(S_{\text{after}})$}, ylabel= { $\log(var)$}]{
		
		\addplot [red, only marks]  table [x=logs, y=logvar]   {\tablethreefive};
		\addplot [green, only marks]  table [x=logs, y=logvar]   {\tablethreexfive};
		\addplot [blue, only marks]  table [x=logs, y=logvar]   {\tablethreehnfive};
		\addplot [magenta, only marks]  table [x=logs, y=logvar]   {\tablethreebnfive};
		\addplot [yellow, only marks]  table [x=logs, y=logvar]   {\tablethreedfive};
		\addplot [olive, solid, domain=-6:24]{(ln(1.25/0.01))+x};
		
		\addplot [red, only marks]  table [x=logs, y=logvar]   {\tablefourfive};
		\addplot [green, only marks]  table [x=logs, y=logvar]   {\tablefourxfive};
		\addplot [blue, only marks]  table [x=logs, y=logvar]   {\tablefourhnfive};
		\addplot [magenta, only marks]  table [x=logs, y=logvar]   {\tablefourbnfive};

		\legend{3-4 layer FC SN, 3-4 layer FC XU, 3-4 layer FC HN, 3-4 layer FC SN BN, 3 layer FC SN dropout, Equation (\ref{eq:varsen})}}
	
	\node[scale=0.75] at (axis cs: 0,16) {$\rho=0.9537$};
	
	\end{axis}
	
	\end{tikzpicture}
	
}
	\caption{$var$ (Equation (\ref{eq:defvar})) versus $S$ (Equation (\ref{eq:sen})) for networks trained on 1000 samples of the CIFAR-10 training dataset for different input $x$ and input noise $\varepsilon_x$ scales. The expression $\Sigma$ is neglected in the computation of (\ref{eq:varsen}) in the figures. \textbf{(a), (b)} The non-normalized original CIFAR-10 input images. \textbf{(c), (d)} Normalized input images with zero-mean and unit variance. \textbf{(e)} Normalized inputs with unit variance and the same mean as the original images. }
	\label{fig:varsen}
\end{figure}

	\section{MNIST AND CIFAR-100 Experiments}\label{app:mnist}
	
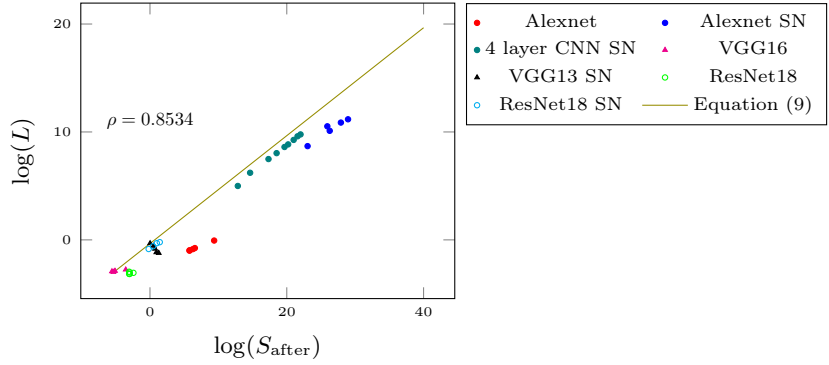
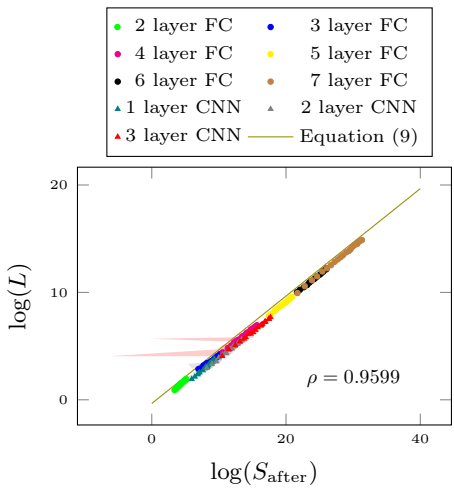
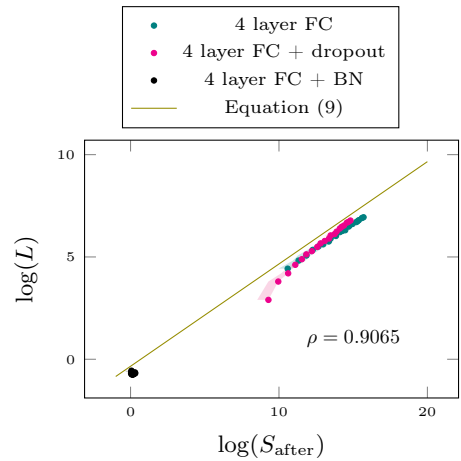
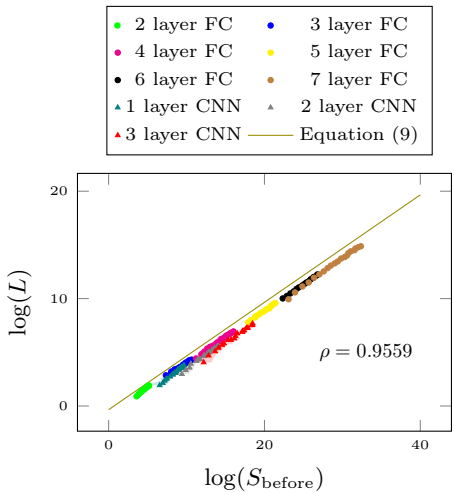
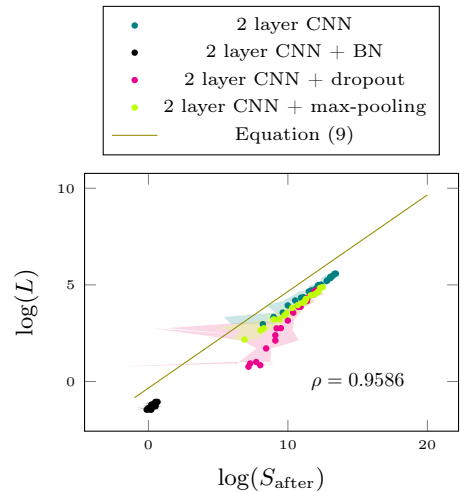
\begin{figure}[h]
		\subfloat[Replicate of Fig. \ref{fig:state} for MNIST]{
			\begin{tikzpicture}[every mark/.append style={mark size=1pt}]
			
			\begin{axis}[width=6.5cm, height=5.5cm, legend pos=outer north east, legend columns = 2, legend style={fill=none, font=\scriptsize}, xlabel= {$\log(S_{\text{after}})$}, ylabel= { $\log(L)$}]{
				
				\addplot [red, only marks]  table [x=logs, y=logloss]   {\tablee};
				\addplot [blue, only marks]  table [x=logs, y=logloss]   {\tabled};
				\addplot [teal, only marks]  table [x=logs, y=logloss]   {\tablea};
				\addplot [magenta, only marks, mark=triangle*]  table [x=logs, y=logloss]   {\tablec};
				\addplot [black, only marks, mark=triangle*]  table [x=logs, y=logloss]   {\tableb};
				\addplot [green, only marks, mark=o]  table [x=logs, y=logloss]   {\tablef};
				\addplot [cyan, only marks, mark=o]  table [x=logs, y=logloss]   {\tableg};
				
				\addplot [olive, solid, domain=-5:40]{0.5*(ln(0.5*0.9*0.112/0.1))+0.5*x};

				\legend{Alexnet, Alexnet SN, 4 layer CNN SN, VGG16, VGG13 SN, ResNet18, ResNet18 SN, Equation (\ref{eq:main})}}
				
			\node[scale=0.7] at (axis cs: 0,11) {$\rho=0.8534$};
			
			\end{axis}
			
			\end{tikzpicture}
			
		}\\
		\subfloat[Fully-connected neural networks]{
		\begin{tikzpicture}[every mark/.append style={mark size=1pt}]
		
		\begin{axis}[width=6.5cm, height=5cm, legend columns = 2, legend style={fill=none, font=\scriptsize, at={(0.95,1.62)}}, xlabel= {$\log(S_{\textrm{after}})$}, ylabel= { $\log(L)$}]

		\addplot [stack plots=x, fill=none, draw=none, forget plot]   table [x=logl, y=logloss]   {\tableone} \closedcycle;
		\addplot [stack plots=x, fill=green!50, opacity=0.4, draw opacity=0, area legend, forget plot]   table [x expr=\thisrow{logu}-\thisrow{logl}, y=logloss]   {\tableone} \closedcycle;
		\addplot [stack plots=x, stack dir=minus, forget plot, draw=none] table [x=logu, y=logloss] {\tableone};
		
		\addplot [stack plots=x, fill=none, draw=none, forget plot]   table [x=logl, y=logloss]   {\tabletwo} \closedcycle;
		\addplot [stack plots=x, fill=blue!50, opacity=0.4, draw opacity=0, area legend, forget plot]   table [x expr=\thisrow{logu}-\thisrow{logl}, y=logloss]   {\tabletwo} \closedcycle;
		\addplot [stack plots=x, stack dir=minus, forget plot, draw=none] table [x=logu, y=logloss] {\tabletwo};
		
		\addplot [stack plots=x, fill=none, draw=none, forget plot]   table [x=logl, y=logloss]   {\tablethree} \closedcycle;
		\addplot [stack plots=x, fill=magenta!50, opacity=0.4, draw opacity=0, area legend, forget plot]   table [x expr=\thisrow{logu}-\thisrow{logl}, y=logloss]   {\tablethree} \closedcycle;
		\addplot [stack plots=x, stack dir=minus, forget plot, draw=none] table [x=logu, y=logloss] {\tablethree};
		\addplot [stack plots=x, fill=none, draw=none, forget plot]   table [x=logl, y=logloss]   {\tablesix} \closedcycle;
		\addplot [stack plots=x, fill=brown!50, opacity=0.4, draw opacity=0, area legend, forget plot]   table [x expr=\thisrow{logu}-\thisrow{logl}, y=logloss]   {\tablesix} \closedcycle;
		\addplot [stack plots=x, stack dir=minus, forget plot, draw=none] table [x=logu, y=logloss] {\tablesix};
		
		\addplot [stack plots=x, fill=none, draw=none, forget plot]   table [x=logl, y=logloss]   {\tableonec} \closedcycle;
		\addplot [stack plots=x, fill=teal!50, opacity=0.4, draw opacity=0, area legend, forget plot]   table [x expr=\thisrow{logu}-\thisrow{logl}, y=logloss]   {\tableonec} \closedcycle;
		\addplot [stack plots=x, stack dir=minus, forget plot, draw=none] table [x=logu, y=logloss] {\tableonec};
		
		\addplot [stack plots=x, fill=none, draw=none, forget plot]   table [x=logl, y=logloss]   {\tabletwoc} \closedcycle;
		\addplot [stack plots=x, fill=gray!50, opacity=0.4, draw opacity=0, area legend, forget plot]   table [x expr=\thisrow{logu}-\thisrow{logl}, y=logloss]   {\tabletwoc} \closedcycle;
		\addplot [stack plots=x, stack dir=minus, forget plot, draw=none] table [x=logu, y=logloss] {\tabletwoc};
		
		\addplot [stack plots=x, fill=none, draw=none, forget plot]   table [x=logl, y=logloss]   {\tablethreec} \closedcycle;
		\addplot [stack plots=x, fill=red!50, opacity=0.4, draw opacity=0, area legend, forget plot]   table [x expr=\thisrow{logu}-\thisrow{logl}, y=logloss]   {\tablethreec} \closedcycle;
		\addplot [stack plots=x, stack dir=minus, forget plot, draw=none] table [x=logu, y=logloss] {\tablethreec};
		
		\addplot [green, only marks]  table [x=logm, y=logloss]   {\tableone};
		\addplot [blue, only marks]  table [x=logm, y=logloss]   {\tabletwo};
		\addplot [magenta, only marks]  table [x=logm, y=logloss]   {\tablethree};
		\addplot [yellow, only marks]  table [x=logm, y=logloss]   {\tablefour};
		\addplot [black, only marks]  table [x=logm, y=logloss]   {\tablefive};
		\addplot [brown, only marks]  table [x=logm, y=logloss]   {\tablesix};
		\addplot [teal, only marks, mark=triangle*]  table [x=logm, y=logloss]   {\tableonec};
		\addplot [gray, only marks,mark=triangle*]  table [x=logm, y=logloss]   {\tabletwoc};
		\addplot [red, only marks, mark=triangle*]  table [x=logm, y=logloss]   {\tablethreec};
		\addplot [olive, solid, domain=0:40]{0.5*(ln(0.5*0.9*0.112/0.1))+0.5*x};


		\node[scale=0.75] at (axis cs: 30,2) {$\rho=0.9599$};

		\legend{2 layer FC, 3 layer FC, 4 layer FC, 5 layer FC, 6 layer FC, 7 layer FC, 1 layer CNN, 2 layer CNN, 3 layer CNN, Equation (\ref{eq:main})}

		\end{axis}
		
		\end{tikzpicture}
	
	}\qquad \qquad
		\subfloat[4-layer fully-connected neural networks trained with or without regularization]{
		\begin{tikzpicture}[every mark/.append style={mark size=1pt}]
		
		
		\begin{axis}[width=6.5cm, height=5cm, legend columns = 1, legend pos=outer north east, legend style={fill=none, font=\scriptsize, at={(0.1,1.52)}}, xlabel= {$\log(S_{\textrm{after}})$}, ylabel= {$\log(L)$}]{
			\addplot [stack plots=x, fill=none, draw=none, forget plot]   table [x=logl, y=logloss]   {\tablethree} \closedcycle;
			\addplot [stack plots=x, fill=teal!50, opacity=0.4, draw opacity=0, area legend, forget plot]   table [x expr=\thisrow{logu}-\thisrow{logl}, y=logloss]   {\tablethree} \closedcycle;
			\addplot [stack plots=x, stack dir=minus, forget plot, draw=none] table [x=logu, y=logloss] {\tablethree};
			
			\addplot [stack plots=x, fill=none, draw=none, forget plot]   table [x=logl, y=logloss]   {\tablethreed} \closedcycle;
			\addplot [stack plots=x, fill=magenta!50, opacity=0.4, draw opacity=0, area legend, forget plot]   table [x expr=\thisrow{logu}-\thisrow{logl}, y=logloss]   {\tablethreed} \closedcycle;
			\addplot [stack plots=x, stack dir=minus, forget plot, draw=none] table [x=logu, y=logloss] {\tablethreed};
			
			\addplot [stack plots=x, fill=none, draw=none, forget plot]   table [x=logl, y=logloss]   {\tablethreebn} \closedcycle;
			\addplot [stack plots=x, fill=black!50, opacity=0.4, draw opacity=0, area legend, forget plot]   table [x expr=\thisrow{logu}-\thisrow{logl}, y=logloss]   {\tablethreebn} \closedcycle;
			\addplot [stack plots=x, stack dir=minus, forget plot, draw=none] table [x=logu, y=logloss] {\tablethreebn};

			\addplot [teal, only marks]  table [x=logm, y=logloss]   {\tablethree};
			\addplot [magenta, only marks]  table [x=logm, y=logloss]   {\tablethreed};
			\addplot [black, only marks]  table [x=logm, y=logloss]   {\tablethreebn};
			\addplot [olive, solid, domain=-1:20]{0.5*(ln(0.5*0.9*0.112/0.1))+0.5*x};
			
			\legend{4 layer FC,4 layer FC + dropout, 4 layer FC + BN, Equation (\ref{eq:main})}

			\node[scale=0.75] at (axis cs: 15,1) {$\rho=0.9065$};}

		\end{axis}

		\end{tikzpicture}
		
		}\\
\subfloat[Sensitivity before training]{
	\begin{tikzpicture}[every mark/.append style={mark size=1pt}]
	
	\begin{axis}[width=6.5cm, height=5cm, legend columns =2, legend style={fill=none, font=\scriptsize, at={(0.95,1.65)}}, xlabel= {$\log(S_{\textrm{before}})$}, ylabel= { $\log(L)$}]

	\addplot [stack plots=x, fill=none, draw=none, forget plot]   table [x=loglb, y=logloss]   {\tableone} \closedcycle;
	\addplot [stack plots=x, fill=green!50, opacity=0.4, draw opacity=0, area legend, forget plot]   table [x expr=\thisrow{logub}-\thisrow{loglb}, y=logloss]   {\tableone} \closedcycle;
	\addplot [stack plots=x, stack dir=minus, forget plot, draw=none] table [x=logub, y=logloss] {\tableone};
	
	\addplot [stack plots=x, fill=none, draw=none, forget plot]   table [x=loglb, y=logloss]   {\tabletwo} \closedcycle;
	\addplot [stack plots=x, fill=blue!50, opacity=0.4, draw opacity=0, area legend, forget plot]   table [x expr=\thisrow{logub}-\thisrow{loglb}, y=logloss]   {\tabletwo} \closedcycle;
	\addplot [stack plots=x, stack dir=minus, forget plot, draw=none] table [x=logub, y=logloss] {\tabletwo};
	
	\addplot [stack plots=x, fill=none, draw=none, forget plot]   table [x=loglb, y=logloss]   {\tablethree} \closedcycle;
	\addplot [stack plots=x, fill=magenta!50, opacity=0.4, draw opacity=0, area legend, forget plot]   table [x expr=\thisrow{logub}-\thisrow{loglb}, y=logloss]   {\tablethree} \closedcycle;
	\addplot [stack plots=x, stack dir=minus, forget plot, draw=none] table [x=logub, y=logloss] {\tablethree};
	\addplot [stack plots=x, fill=none, draw=none, forget plot]   table [x=loglb, y=logloss]   {\tablesix} \closedcycle;
	\addplot [stack plots=x, fill=brown!50, opacity=0.4, draw opacity=0, area legend, forget plot]   table [x expr=\thisrow{logub}-\thisrow{loglb}, y=logloss]   {\tablesix} \closedcycle;
	\addplot [stack plots=x, stack dir=minus, forget plot, draw=none] table [x=logub, y=logloss] {\tablesix};

	\addplot [stack plots=x, fill=none, draw=none, forget plot]   table [x=loglb, y=logloss]   {\tableonec} \closedcycle;
	\addplot [stack plots=x, fill=teal!50, opacity=0.4, draw opacity=0, area legend, forget plot]   table [x expr=\thisrow{logub}-\thisrow{loglb}, y=logloss]   {\tableonec} \closedcycle;
	\addplot [stack plots=x, stack dir=minus, forget plot, draw=none] table [x=logub, y=logloss] {\tableonec};
	
	\addplot [stack plots=x, fill=none, draw=none, forget plot]   table [x=loglb, y=logloss]   {\tabletwoc} \closedcycle;
	\addplot [stack plots=x, fill=gray!50, opacity=0.4, draw opacity=0, area legend, forget plot]   table [x expr=\thisrow{logub}-\thisrow{loglb}, y=logloss]   {\tabletwoc} \closedcycle;
	\addplot [stack plots=x, stack dir=minus, forget plot, draw=none] table [x=logub, y=logloss] {\tabletwoc};
	
	\addplot [stack plots=x, fill=none, draw=none, forget plot]   table [x=loglb, y=logloss]   {\tablethreec} \closedcycle;
	\addplot [stack plots=x, fill=red!50, opacity=0.4, draw opacity=0, area legend, forget plot]   table [x expr=\thisrow{logub}-\thisrow{loglb}, y=logloss]   {\tablethreec} \closedcycle;
	\addplot [stack plots=x, stack dir=minus, forget plot, draw=none] table [x=logub, y=logloss] {\tablethreec};

	\addplot [green, only marks]  table [x=logmb, y=logloss]   {\tableone};
	\addplot [blue, only marks]  table [x=logmb, y=logloss]   {\tabletwo};
	\addplot [magenta, only marks]  table [x=logmb, y=logloss]   {\tablethree};
	\addplot [yellow, only marks]  table [x=logmb, y=logloss]   {\tablefour};
	\addplot [black, only marks]  table [x=logmb, y=logloss]   {\tablefive};
	\addplot [brown, only marks]  table [x=logmb, y=logloss]   {\tablesix};
	\addplot [teal, only marks, mark=triangle*]  table [x=logmb, y=logloss]   {\tableonec};
	\addplot [gray, only marks, mark=triangle*]  table [x=logmb, y=logloss]   {\tabletwoc};
	\addplot [red, only marks, mark=triangle*]  table [x=logmb, y=logloss]   {\tablethreec};
	
	\addplot [olive, solid, domain=0:40]{0.5*(ln(0.5*0.9*0.112/0.1))+0.5*x};

	\node[scale=0.75] at (axis cs: 33,5) {$\rho=0.9559$};
	
	\legend{2 layer FC, 3 layer FC, 4 layer FC, 5 layer FC, 6 layer FC, 7 layer FC, 1 layer CNN, 2 layer CNN, 3 layer CNN, Equation (\ref{eq:main})}
	
	\end{axis}
	
	\end{tikzpicture}
	
}\qquad \qquad
		\subfloat[2-layer convolutional neural networks trained with or without regularization techniques]{
		\begin{tikzpicture}[every mark/.append style={mark size=1pt}]
		
		\begin{axis}[width=6.5cm, height=5cm, legend columns = 1,legend style={fill=none, font=\scriptsize, at={(0.95,1.65)}}, xlabel= {$\log(S_{\textrm{after}})$}, ylabel= {$\log(L)$}]{
			
			\addplot [stack plots=x, fill=none, draw=none, forget plot]   table [x=logl, y=logloss]   {\tabletwoc} \closedcycle;
			\addplot [stack plots=x, fill=teal!50, opacity=0.4, draw opacity=0, area legend, forget plot]   table [x expr=\thisrow{logu}-\thisrow{logl}, y=logloss]   {\tabletwoc} \closedcycle;
			\addplot [stack plots=x, stack dir=minus, forget plot, draw=none] table [x=logu, y=logloss] {\tabletwoc};
			
			\addplot [stack plots=x, fill=none, draw=none, forget plot]   table [x=logl, y=logloss]   {\tabletwobnc} \closedcycle;
			\addplot [stack plots=x, fill=black!50, opacity=0.4, draw opacity=0, area legend, forget plot]   table [x expr=\thisrow{logu}-\thisrow{logl}, y=logloss]   {\tabletwobnc} \closedcycle;
			\addplot [stack plots=x, stack dir=minus, forget plot, draw=none] table [x=logu, y=logloss] {\tabletwobnc};
			
			\addplot [stack plots=x, fill=none, draw=none, forget plot]   table [x=logl, y=logloss]   {\tabletwodc} \closedcycle;
			\addplot [stack plots=x, fill=magenta!50, opacity=0.4, draw opacity=0, area legend, forget plot]   table [x expr=\thisrow{logu}-\thisrow{logl}, y=logloss]   {\tabletwodc} \closedcycle;
			\addplot [stack plots=x, stack dir=minus, forget plot, draw=none] table [x=logu, y=logloss] {\tabletwodc};
			
			\addplot [stack plots=x, fill=none, draw=none, forget plot]   table [x=logl, y=logloss]   {\tabletwomaxc} \closedcycle;
			\addplot [stack plots=x, fill=lime!50, opacity=0.4, draw opacity=0, area legend, forget plot]   table [x expr=\thisrow{logu}-\thisrow{logl}, y=logloss]   {\tabletwomaxc} \closedcycle;
			\addplot [stack plots=x, stack dir=minus, forget plot, draw=none] table [x=logu, y=logloss] {\tabletwomaxc};

			\addplot [teal, only marks]  table [x=logm, y=logloss]   {\tabletwoc};
			\addplot [black, only marks]  table [x=logm, y=logloss]   {\tabletwobnc};
			\addplot [magenta, only marks]  table [x=logm, y=logloss]   {\tabletwodc};
			\addplot [lime, only marks]  table [x=logm, y=logloss]   {\tabletwomaxc};
			\addplot [olive, solid, domain=-1:20]{0.5*(ln(0.5*0.9*0.112/0.1))+0.5*x};
			
			\legend{2 layer CNN, 2 layer CNN + BN, 2 layer CNN + dropout, 2 layer CNN + max-pooling, Equation (\ref{eq:main})}

		\node[scale=0.75] at (axis cs: 15,0) {$\rho=0.9586$};}
		
		\end{axis}
		
		\end{tikzpicture}
	}\\

	\caption{Test loss $S$ versus sensitivity $S$ for networks trained on 6000 samples of the MNIST training dataset. Each point in each color indicates a network with a different width and the sensitivity and test loss are averaged over multiple runs.}
	\label{fig:fcnn_mnist}
	\end{figure}

\begin{figure}[h]
		\subfloat[Replicate of Fig. \ref{fig:state} for CIFAR-100]{
			\begin{tikzpicture}[every mark/.append style={mark size=1pt}]
			
			\begin{axis}[width=6.5cm, height=5.5cm, legend columns = 2, legend pos=outer north east, legend style={font=\scriptsize}, xlabel= {$\log(S_{\text{after}})$}, ylabel= { $\log(L)$}]{
				
				\addplot [red, only marks]  table [x=logs, y=logloss]   {\tablee};
				\addplot [blue, only marks]  table [x=logs, y=logloss]   {\tabled};
				\addplot [teal, only marks]  table [x=logs, y=logloss]   {\tablea};
				\addplot [magenta, only marks]  table [x=logs, y=logloss]   {\tablec};
				\addplot [black, only marks]  table [x=logs, y=logloss]   {\tableb};
				\addplot [green, only marks]  table [x=logs, y=logloss]   {\tablef};
				\addplot [cyan, only marks]  table [x=logs, y=logloss]   {\tableg};
					
				\addplot [olive, solid, domain=-5:40]{0.5*(ln(0.5*0.99*0.2648/0.01))+0.5*x};

				\legend{Alexnet, Alexnet SN, 4 layer CNN SN, VGG16, VGG13 SN, ResNet18, ResNet18 SN, Equation (\ref{eq:main})}}
			
			\node[scale=0.75] at (axis cs: 1,11) {$\rho=0.9695$};
			
			\end{axis}
			
			\end{tikzpicture}
			
		}\\
		\subfloat[Sensitivity after training vs test loss]{
			\begin{tikzpicture}[every mark/.append style={mark size=1pt}]
			
			\begin{axis}[width=6.5cm, height=5.5cm, legend pos=outer north east, legend style={font=\scriptsize}, xlabel= {$\log(S_{\text{after}})$}, ylabel= { $\log(L)$}]{
				
				\addplot [green, only marks]  table [x=logs, y=logloss]   {\tabletwo};
				\addplot [blue, only marks]  table [x=logs, y=logloss]   {\tablethree};
				\addplot [magenta, only marks]  table [x=logs, y=logloss]   {\tablefour};
				\addplot [yellow, only marks]  table [x=logs, y=logloss]   {\tablefive};
				\addplot [black, only marks]  table [x=logs, y=logloss]   {\tablesix};
				\addplot [brown, only marks]  table [x=logs, y=logloss]   {\tableseven};
				\addplot [teal, only marks, mark=triangle*]  table [x=logs, y=logloss]   {\tableonec};
				\addplot [gray, only marks, mark=triangle*]  table [x=logs, y=logloss]   {\tabletwoc};
				\addplot [red, only marks, mark=triangle*]  table [x=logs, y=logloss]   {\tablethreec};
				\addplot [olive, solid, domain=5:48]{0.5*(ln(0.5*0.99*0.2648/0.01))+0.5*x};

				\legend{2 layer FC, 3 layer FC, 4 layer FC, 5 layer FC, 6 layer FC, 7 layer FC, 1 layer CNN, 2 layer CNN, 3 layer CNN, Equation (\ref{eq:main})}}
			
			\node[scale=0.75] at (axis cs: 11,20) {$\rho=0.9708$};
			
			\end{axis}
			
			\end{tikzpicture}
			
		}\qquad
	\subfloat[Effect of regularization on fully-connected neural networks]{
		\begin{tikzpicture}[every mark/.append style={mark size=1pt}]
		
		\begin{axis}[width=6.6cm, height=5.5cm, legend pos=north west, legend style={fill=none, font=\scriptsize}, xlabel= {$\log(S_{\text{after}})$}, ylabel= { $\log(L)$}]{
			
			\addplot [red, only marks]  table [x=logs, y=logloss]   {\tableseven};
			\addplot [blue, only marks]  table [x=logs, y=logloss]   {\tablesevend};
			\addplot [green, only marks]  table [x=logs, y=logloss]   {\tablesevenbn};
			\addplot [olive, solid, domain=0:47]{0.5*(ln(0.5*0.99*0.2648/0.01))+0.5*x};

			\legend{ 7 layer FC, 7 layer FC + dropout, 7 layer FC + BN, Equation (\ref{eq:main})}}
		
		\node[scale=0.75] at (axis cs: 38,5) {$\rho=0.9524$};
		
		\end{axis}
		
		\end{tikzpicture}
		
	}\\
\subfloat[Sensitivity before training vs test loss]{
	\begin{tikzpicture}[every mark/.append style={mark size=1pt}]
	
	\begin{axis}[width=6.5cm, height=5.5cm, legend pos=outer north east, legend style={fill=none, font=\scriptsize}, xlabel= {$\log(S_{\text{before}})$}, ylabel= { $\log(L)$}]{
		
		\addplot [green, only marks]  table [x=logsb, y=logloss]   {\tabletwo};
		\addplot [blue, only marks]  table [x=logsb, y=logloss]   {\tablethree};
		\addplot [magenta, only marks]  table [x=logsb, y=logloss]   {\tablefour};
		\addplot [yellow, only marks]  table [x=logsb, y=logloss]   {\tablefive};
		\addplot [black, only marks]  table [x=logsb, y=logloss]   {\tablesix};
		\addplot [brown, only marks]  table [x=logsb, y=logloss]   {\tableseven};
		\addplot [teal, only marks, mark=triangle*]  table [x=logsb, y=logloss]   {\tableonec};
		\addplot [gray, only marks, mark=triangle*]  table [x=logsb, y=logloss]   {\tabletwoc};
		\addplot [red, only marks, mark=triangle*]  table [x=logsb, y=logloss]   {\tablethreec};
		\addplot [olive, solid, domain=-5:48]{0.5*(ln(0.5*0.99*0.2648/0.01))+0.5*x};

		\legend{2 layer FC, 3 layer FC, 4 layer FC, 5 layer FC, 6 layer FC, 7 layer FC, 1 layer CNN, 2 layer CNN, 3 layer CNN, Equation (\ref{eq:main})}}
	
	\node[scale=0.75] at (axis cs: 1,11) {$\rho=0.9277$};
	
	\end{axis}
	
	\end{tikzpicture}
	
}\qquad
\subfloat[Effect of regularization on convolutional neural networks]{
	\begin{tikzpicture}[every mark/.append style={mark size=1pt}]
	
	\begin{axis}[width=6.6cm, height=5.5cm, legend pos=north west, legend style={fill=none, font=\scriptsize}, xlabel= {$\log(S_{\text{after}})$}, ylabel= { $\log(L)$}]{
		
		\addplot [red, only marks]  table [x=logs, y=logloss]   {\tablethreec};
		\addplot [blue, only marks]  table [x=logs, y=logloss]   {\tablethreecd};
		\addplot [green, only marks]  table [x=logs, y=logloss]   {\tablethreecbn};
		\addplot [olive, solid, domain=0:30]{0.5*(ln(0.5*0.99*0.2648/0.01))+0.5*x};

		\legend{ 3 layer CNN, 3 layer CNN + dropout, 3 layer CNN + BN, Equation (\ref{eq:main})}}
	
	\node[scale=0.75] at (axis cs: 24,4) {$\rho=0.9644$};
	
	\end{axis}
	
	\end{tikzpicture}
	
}

	\caption{Test loss $L$ versus sensitivity $S$ for networks trained on 1000 samples of the CIFAR-100 training dataset. Each point indicates a network with a different width and the sensitivity and test loss are averaged over 10 runs.}
\label{fig:cifar100}
\end{figure}
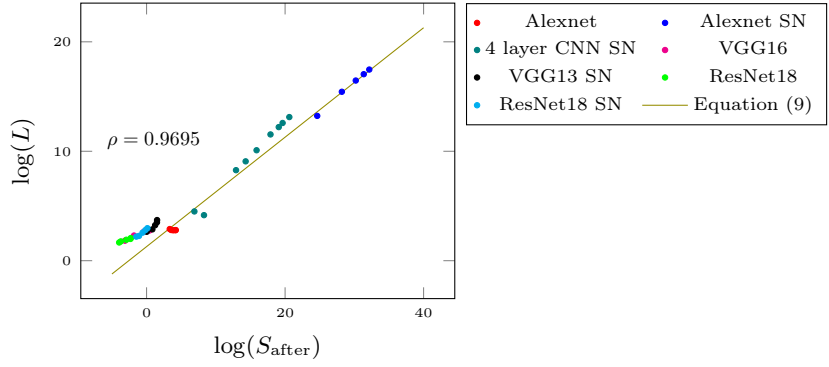
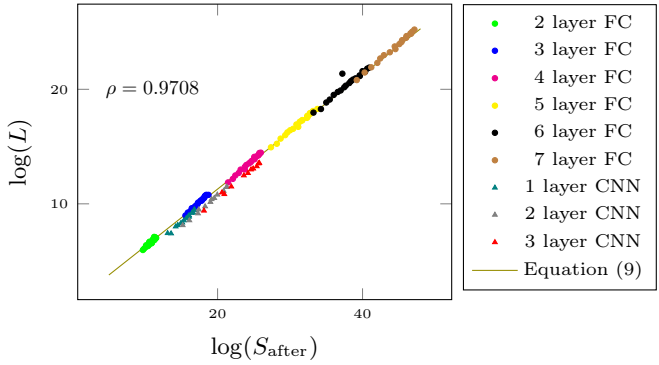
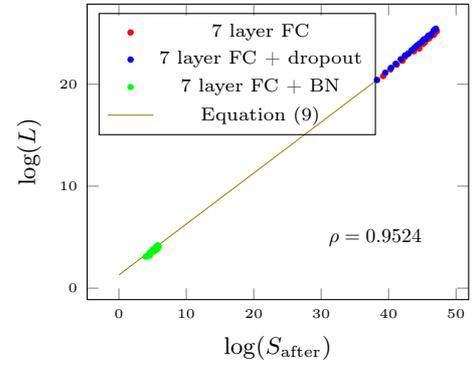
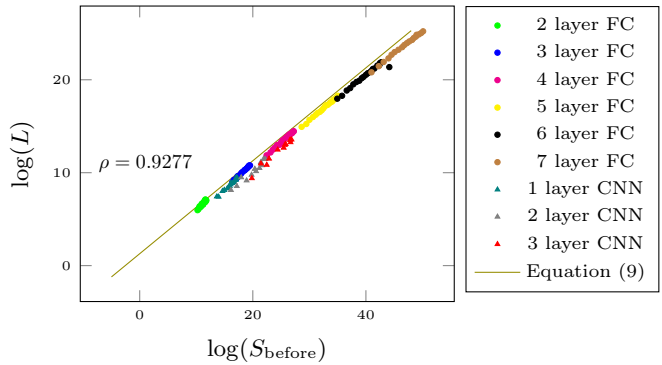
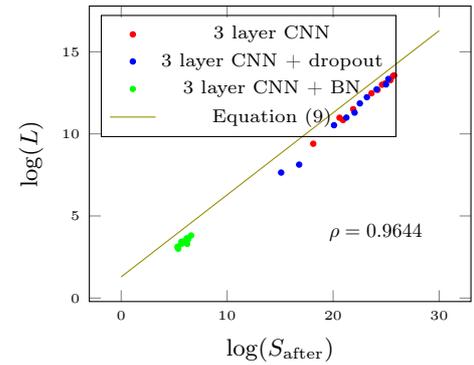

	In this section, we present the experimental results for networks trained on 6000 samples of the MNIST\footnote{\url{http://yann.lecun.com/exdb/mnist/}} training dataset and evaluated on the entire MNIST testing dataset. Figs.~\ref{fig:fcnn_mnist}~(a) and~(b) show the results for fully-connected neural networks with different numbers of layers and hidden units and using regularization techniques batch normalization and dropout. Figs.~\ref{fig:fcnn_mnist}~(c) and~(d) show the results for convolutional neural networks. Finally, Figs.~\ref{fig:fcnn_mnist}~(e) and~(f) show the results on the comparison of the sensitivity of untrained networks $S_\text{before}$ with the test loss $L$ after the networks are trained.
	Fig.~\ref{fig:cifar100} shows the sensitivity $S$ versus the loss $L$ for networks trained on 1000 samples of the CIFAR-100 dataset\footnote{\url{https://www.cs.toronto.edu/~kriz/cifar.html}}. The empirical results on these two datasets also show a rather strong match to~(\ref{eq:main}), and once again we observe the relation between sensitivity and generalization and the effect of state-of-the-art techniques on both sensitivity and generalization.

	\clearpage
	
	\begin{table}[]
		\caption{Comparison of Pearson's correlation coefficient $\rho$ between cross-entropy loss $L$, sensitivity $S$ and classification error for 4-layer CNN and 4-layer FC trained on a subset of the MNIST dataset before and after applying temperature scaling~\cite{guo2017calibration}, which is a network calibration method.}\label{tab:cal}
		\begin{tabular}{|c|c|c|}
			\hline
			& before calibration & after calibration \\
			\hline
			$\rho$ between $L$ and $S$                  & 0.958              & 0.841             \\
			\hline
			$\rho$ between $L$ and classification error & -0.797             & 0.137             \\
			\hline
			$\rho$ between $S$ and classification error & -0.757             & 0.087            \\
			\hline
		\end{tabular}
	\end{table}

	\begin{figure*}[h]
		\subfloat[Test loss versus variance]{
		\begin{tikzpicture}[every mark/.append style={mark size=1pt}]
		
		\begin{axis}[width=6cm, height=5cm, legend pos=outer north east, legend style={font=\scriptsize}, xlabel= {$\log(\varepsilon_{variance})$}, ylabel= { $\log(L_\text{MSE})$}]

		\addplot [red, only marks]  table [x=logm, y=logloss]   {\tableone};
		\addplot [blue, only marks]  table [x=logm, y=logloss]   {\tabletwo};
		\addplot [green, only marks]  table [x=logm, y=logloss]   {\tablethree};
		\addplot [cyan, only marks]  table [x=logm, y=logloss]   {\tablefour};
		\addplot [magenta, only marks]  table [x=logm, y=logloss]   {\tablefive};
		\addplot [black, only marks]  table [x=logm, y=logloss]   {\tablesix};
		
		\legend{3 layer FC, 4 layer FC, 5 layer FC, 6 layer FC, 7 layer FC, 8 layer FC}
		
		\end{axis}
		
		\end{tikzpicture}
	}\qquad
		\subfloat[Test loss versus bias]{
\begin{tikzpicture}[every mark/.append style={mark size=1pt}]

\begin{axis}[width=6cm, height=5cm, legend pos=outer north east, legend style={font=\scriptsize}, xlabel= {$\log(\varepsilon_{bias})$}, ylabel= {$\log(L_\text{MSE})$}]

	\addplot [red, only marks]  table [x=logm, y=logloss]   {\tableonebias};
	\addplot [blue, only marks]  table [x=logm, y=logloss]   {\tabletwobias};
	\addplot [green, only marks]  table [x=logm, y=logloss]   {\tablethreebias};
	\addplot [cyan, only marks]  table [x=logm, y=logloss]   {\tablefourbias};
	\addplot [magenta, only marks]  table [x=logm, y=logloss]   {\tablefivebias};
	\addplot [black, only marks]  table [x=logm, y=logloss]   {\tablesixbias};
	
	\legend{3 layer FC, 4 layer FC, 5 layer FC, 6 layer FC, 7 layer FC, 8 layer FC}
	
\end{axis}

\end{tikzpicture}

}\\
\subfloat[Test loss versus sum of bias and variance]{

\begin{tikzpicture}[every mark/.append style={mark size=1pt}]

\begin{axis}[width=6cm, height=5cm, legend pos=outer north east, legend style={font=\scriptsize}, xlabel= {$\log(\varepsilon_{bias}+\varepsilon_{variance})$}, ylabel= { $\log(L_\text{MSE})$}]

\addplot [red, only marks]  table [x=logm, y=logloss]   {\tableonep};
\addplot [blue, only marks]  table [x=logm, y=logloss]   {\tabletwop};
\addplot [green, only marks]  table [x=logm, y=logloss]   {\tablethreep};
\addplot [cyan, only marks]  table [x=logm, y=logloss]   {\tablefourp};
\addplot [magenta, only marks]  table [x=logm, y=logloss]   {\tablefivep};
\addplot [black, only marks]  table [x=logm, y=logloss]   {\tablesixp};

\legend{3 layer FC, 4 layer FC, 5 layer FC, 6 layer FC, 7 layer FC, 8 layer FC}

\node[scale=0.75] at (axis cs: 9,35) {$\rho=0.9999$};

\end{axis}

\end{tikzpicture}

}\qquad
\subfloat[Variance versus sensitivity $S_\text{after}$]{
	
	\begin{tikzpicture}[every mark/.append style={mark size=1pt}]
	
	\begin{axis}[width=6cm, height=5cm, legend pos=outer north east, legend style={font=\scriptsize}, xlabel= {$\log(S_\text{after})$}, ylabel= { $\log(\varepsilon_{\textrm{variance}})$}]

	\addplot [red, only marks]  table [x=logms, y=logvar]   {\tableonep};
	\addplot [blue, only marks]  table [x=logms, y=logvar]   {\tabletwop};
	\addplot [green, only marks]  table [x=logms, y=logvar]   {\tablethreep};
	\addplot [cyan, only marks]  table [x=logms, y=logvar]   {\tablefourp};
	\addplot [magenta, only marks]  table [x=logms, y=logvar]   {\tablefivep};
	\addplot [black, only marks]  table [x=logms, y=logvar]   {\tablesixp};
	
	\legend{3 layer FC, 4 layer FC, 5 layer FC, 6 layer FC, 7 layer FC, 8 layer FC}

	\end{axis}
	
	\end{tikzpicture}

}\\
\subfloat[Test loss versus sensitivity $S_\text{after}$]{
\begin{tikzpicture}[every mark/.append style={mark size=1pt}]

\begin{axis}[width=6cm, height=5cm, legend pos=outer north east, legend style={fill= none, font=\scriptsize}, xlabel= {$\log(S_\text{after})$}, ylabel= {$\log(L_{\text{MSE}})$ }]

\addplot [red, only marks]  table [x=logms, y=logloss]   {\tableonep};
\addplot [blue, only marks]  table [x=logms, y=logloss]   {\tabletwop};
\addplot [green, only marks]  table [x=logms, y=logloss]   {\tablethreep};
\addplot [cyan, only marks]  table [x=logms, y=logloss]   {\tablefourp};
\addplot [magenta, only marks]  table [x=logms, y=logloss]   {\tablefivep};
\addplot [black, only marks]  table [x=logms, y=logloss]   {\tablesixp};


\legend{3 layer FC, 4 layer FC, 5 layer FC, 6 layer FC, 7 layer FC, 8 layer FC, theory}

\node[scale=0.75] at (axis cs: 35,8) {$\rho=0.7681$};

\end{axis}

\end{tikzpicture}

}
		\caption{Test loss $L$ versus variance $\varepsilon_{\textrm{variance}}$, bias $\varepsilon_{\textrm{bias}}$ and sensitivity $S$ for a regression task using the MSE loss. The fully-connected neural networks are trained and evaluated on the Boston house price dataset. Each point represents a network with a different width and its coordinates are averaged over multiple runs.}\label{fig:reg2_long}
	\end{figure*}
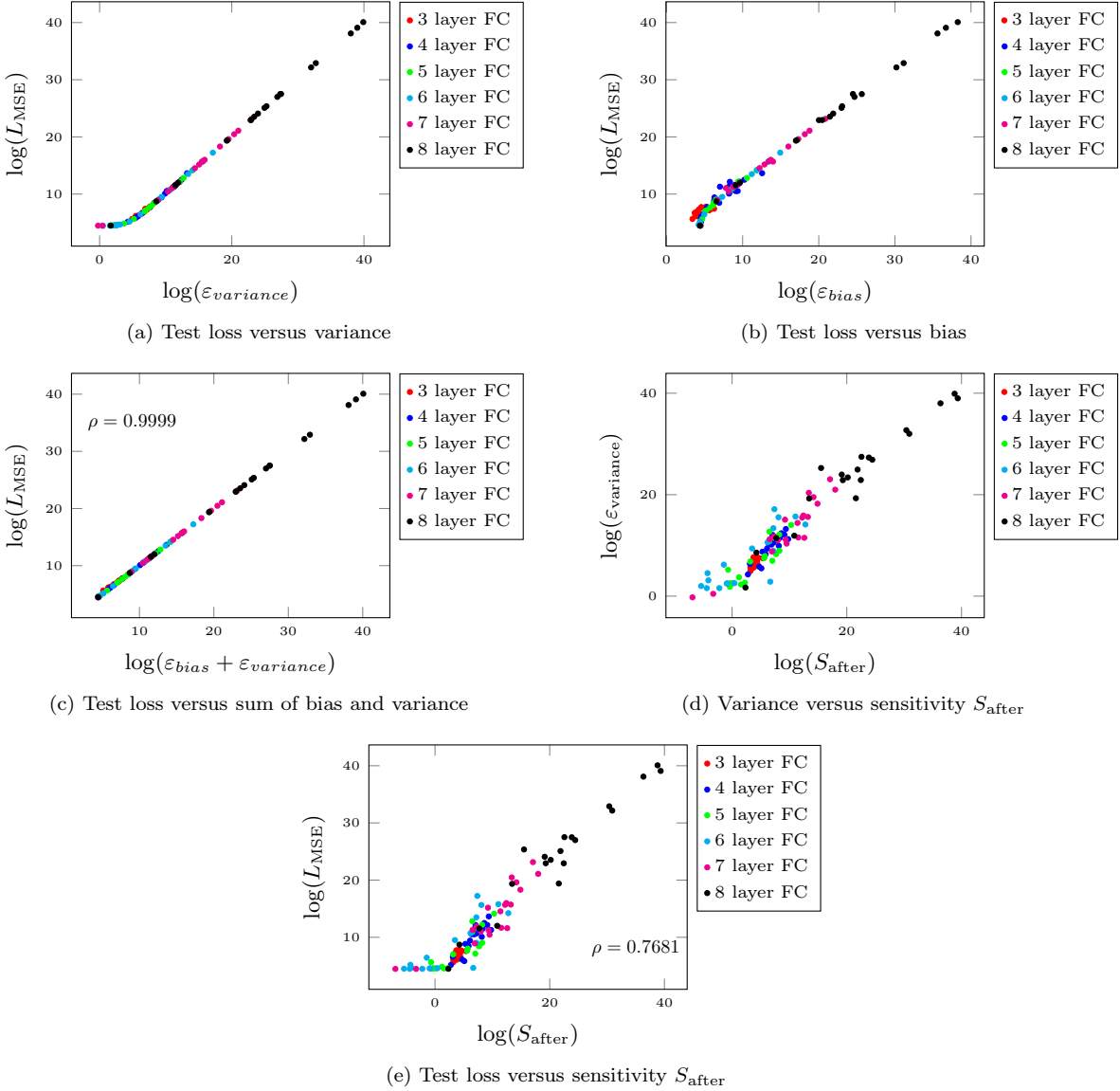

	\begin{figure*}[h]
		\subfloat[Networks trained on 1000 samples of the training dataset]{
		\begin{tikzpicture}[every mark/.append style={mark size=1pt}]
		
		\begin{axis}[width=6.5cm, height=6cm,  legend style={fill=none, font=\scriptsize, at={(1,1.35)}}, xlabel= {$\log(S_\text{after})$}, ylabel= { $\log(L)$}]

		\addplot [black, only marks]  table [x=logm, y=logloss]   {\tableone};
		
		\addplot [red, only marks]  table [x=logm, y=logloss]   {\tableresnetone};
		
		\addplot [olive, solid, domain=-5:15]{0.5*(ln(0.5*0.9*0.2648/0.01))+0.5*x};

		\legend{1-2 layer CNN SN, ResNet18 HN and ResNet34 HN, Equation (\ref{eq:main})}

		\end{axis}

		\end{tikzpicture}
	
}\qquad
	\subfloat[ResNet18 and ResNet34, the yellow marks are ResNet18 with s=1, and the green marks are ResNet34 with s=0.5]{

		\begin{tikzpicture}[every mark/.append style={mark size=1pt}]
		
		\begin{axis}[width=6.5cm, height=6cm, legend style={fill=none, font=\scriptsize, at={(0.9,1.5)}}, xlabel= {$\log(S_\text{after})$}, ylabel= { $\log(L)$}]
		
		\addplot [red, only marks]  table [x=logm, y=logloss]   {\tableresnetone};
		\addplot [blue, only marks]  table [x=logm, y=logloss]   {\tableresnettwo};
		\addplot [teal, only marks]  table [x=logm, y=logloss]   {\tableresnetthree};
		\addplot [magenta, only marks]  table [x=logm, y=logloss]   {\tableresnetfour};
		
		\addplot [olive, solid, domain=-5:0]{0.5*(ln(0.5*0.9*0.2648/0.01))+0.5*x};

		\legend{trained on 1000 samples, trained on 2000 samples, trained on 5000 samples, trained on 10000 samples, Equation (\ref{eq:main})}

		\filldraw [fill=yellow, draw=black, thick] (axis cs:-4.034,1.288) circle [radius=4pt];
		\filldraw [fill=yellow, draw=black, thick] (axis cs:-3.953,1.2) circle [radius=4pt];
		\filldraw [fill=yellow, draw=black, thick] (axis cs:-3.642,0.922) circle [radius=4pt];
		\filldraw [fill=yellow, draw=black, thick] (axis cs:-2.718,0.8725) circle [radius=4pt];
		
		\filldraw [fill=green, draw=black, thick] (axis cs:-2.85,1.465) circle [radius=4pt];
		\filldraw [fill=green, draw=black, thick] (axis cs:-2.867,1.423) circle [radius=4pt];
		\filldraw [fill=green, draw=black, thick] (axis cs:-2.97,1.207) circle [radius=4pt];
		\filldraw [fill=green, draw=black, thick] (axis cs:-2.095,1.0509) circle [radius=4pt];
		
		\end{axis}

		\end{tikzpicture}
}\\
\subfloat[1-2 layer CNN]{

	\begin{tikzpicture}[every mark/.append style={mark size=1pt}]
	
	\begin{axis}[width=6.5cm, height=6cm, legend style={fill=none, font=\scriptsize, at={(0.9,1.5)}}, xlabel= {$\log(S_\text{after})$}, ylabel= { $\log(L)$}]
	
	 
	\addplot [magenta, only marks]  table [x=logm, y=logloss]   {\tableonea};
	\addplot [black, only marks]  table [x=logm, y=logloss]   {\tableone};
	\addplot [teal, only marks]  table [x=logm, y=logloss]   {\tableoneb};
	\addplot [blue, only marks]  table [x=logm, y=logloss]   {\tableonec};
	
	\addplot [olive, solid, domain=4:20]{0.5*(ln(0.5*0.9*0.2648/0.01))+0.5*x};

	\legend{trained on 100 samples, trained on 1000 samples, trained on 2000 samples, trained on 5000 samples, Equation (\ref{eq:main})}

	\end{axis}

	\end{tikzpicture}}\qquad
	\subfloat[1-2 layer CNN]{

		\begin{tikzpicture}[every mark/.append style={mark size=1pt}]
		
		\begin{axis}[width=6.5cm, height=6cm, legend style={fill=none, font=\scriptsize, at={(0.92,1.5)}}, xlabel= {$\log(S)$}, ylabel= { $\log(L)$}]
		
		\addplot [magenta, only marks]  table [x=logm, y=logloss]   {\tablefour};
		\addplot [blue, only marks]  table [x=logm, y=logloss]   {\tablethree};
		\addplot [teal, only marks]  table [x=logm, y=logloss]   {\tabletwo};
		\addplot [black, only marks]  table [x=logm, y=logloss]   {\tableone};
		
		\addplot [olive, solid, domain=4:20]{0.5*(ln(0.5*0.9*0.2648/0.01))+0.5*x};

		\legend{not trained, after 10 epochs of training, after 100 epochs of training, fully trained, Equation (\ref{eq:main})}

		\end{axis}

		\end{tikzpicture}}
		
		\caption{Test loss $L$ versus sensitivity $S$ for networks at different stages of training and trained on different numbers of training samples. Each point indicates an average over multiple runs of a network with a different width and depth. (b) is the zoom in of (a) on the bottom left, and we add the results of the same networks trained on a different number of samples. In (b) the network parameters are initially drawn from a normal distribution by using the He technique. (c) and (d) are the zoom in of (a) on the top right, and we add the results for the same networks trained with different number of training samples in (c) and at different stages of training in (d). In~(c) and (d) the network parameters are drawn from the standard normal distribution. In all the figures the red and black points are the same experiments (1-2 layer CNNs trained on 1000 samples for the black points and ResNet18 and ResNet34 trained on 1000 samples for the red points). In (b) we observe how adding number of training samples, results in closer values to (\ref{eq:main}). In (d) with some abuse of notation, $L$ is computed at different stages of training.}
		\label{fig:bias_long}
		
	\end{figure*}
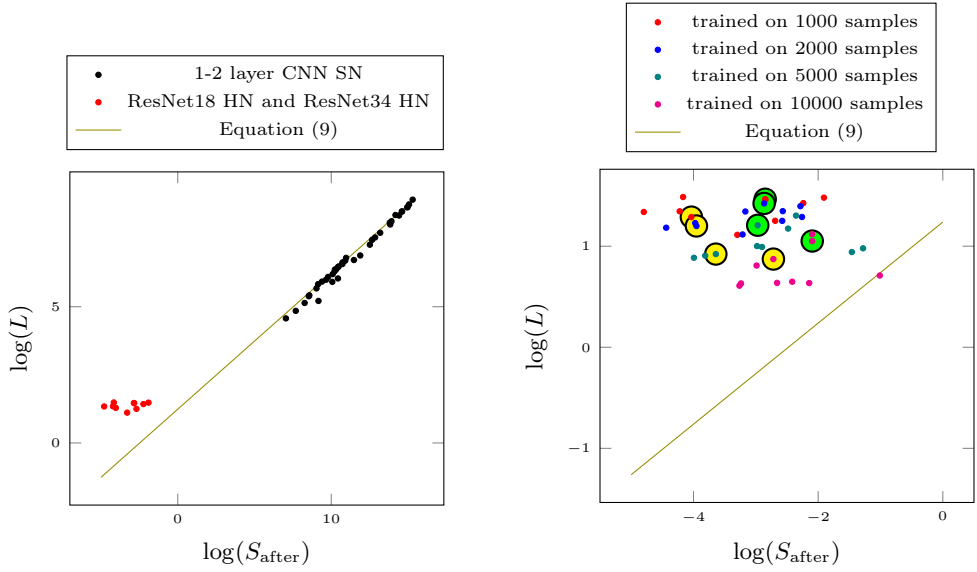
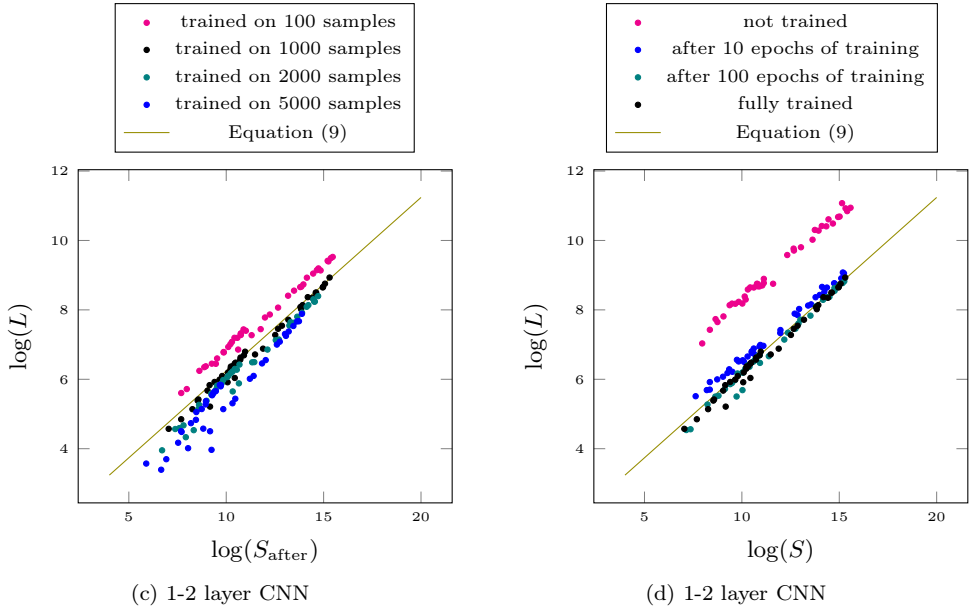

	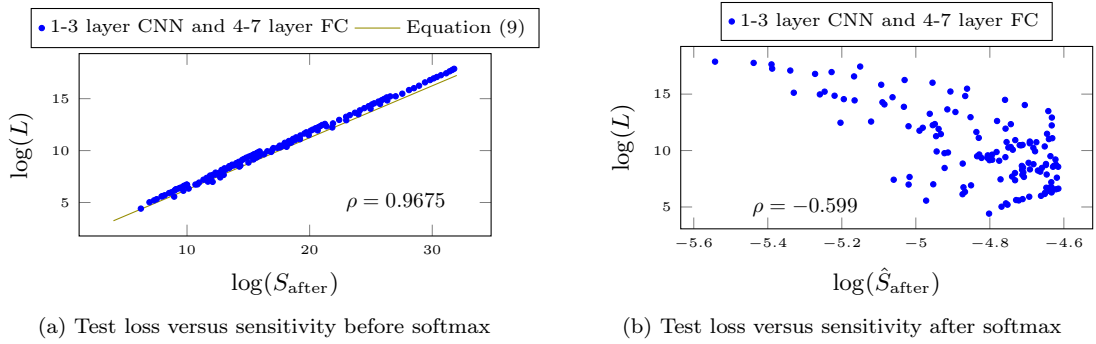
\begin{figure*}[h]
		
\subfloat[Test loss versus sensitivity before softmax]{

	\begin{tikzpicture}[every mark/.append style={mark size=1pt}]
	
	\begin{axis}[width=7cm, height=4cm, legend columns=2, legend style={fill=none, font=\scriptsize, at={(0.5,1.03)},
		anchor=south}, xlabel= {$\log(S_\text{after})$}, ylabel= { $\log(L)$}]
	
	 
	\addplot [blue, only marks]  table [x=logm, y=logloss]   {\tableone};
	
	\addplot [olive, solid, domain=4:32]{0.5*(ln(0.5*0.9*0.2648/0.01))+0.5*x};

	\legend{1-3 layer CNN and 4-7 layer FC, Equation (\ref{eq:main})}
	
	\node[scale=0.8] at (axis cs: 27,5) {$\rho=0.9675$};
	
	\end{axis}

	\end{tikzpicture}}\qquad
	\subfloat[Test loss versus sensitivity after softmax]{

		\begin{tikzpicture}[every mark/.append style={mark size=1pt}]
		
		\begin{axis}[width=7cm, height=4cm, legend columns=2, legend style={fill=none, font=\scriptsize, at={(0.5,1.03)},
			anchor=south}, xlabel= {$\log(\hat{S}_\text{after})$}, ylabel= { $\log(L)$}]
		
		\addplot [blue, only marks]  table [x=logm, y=logloss]   {\tabletwo};
		

		\legend{1-3 layer CNN and 4-7 layer FC}
		
		\node[scale=0.8] at (axis cs: -5.3,5) {$\rho=-0.599$};
		
		\end{axis}

		\end{tikzpicture}}
		
		\caption{Test loss $L$ versus sensitivity before and after the softmax layer for 1-3 layer CNNs and 4-7 layer FCs. The networks are trained on a subset of the CIFAR-10 training set. $S_\text{after}$ is computed after training and \emph{before} the softmax layer and follows~(\ref{eq:sen}); $\hat{S}_\text{after}$ is computed after training and \emph{after} the softmax layer, i.e., $\hat{S}_\text{after}=\E_{\theta^*}\left[ \mathrm{Var}_{x, \varepsilon_x}\left[\frac{1}{K} \sum_{k=1}^{K} F_{\theta^*}^k(x+\varepsilon_x) - F_{\theta^*}^k(x)\right]\right]$. By expanding $\hat{S}_\text{after}$ up to the first order, it is approximated by the product of $\sigma_{\varepsilon_x}^2$ and the Frobenius norm of the Jacobian $J$ of the output. }
		\label{fig:softmax}
		
	\end{figure*}

	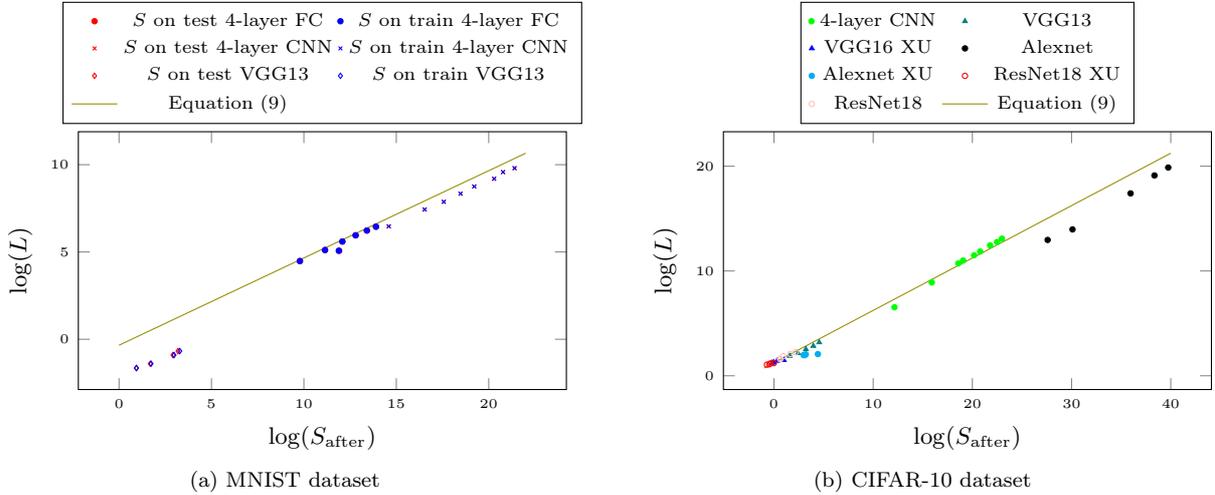
\begin{figure*}[h]
	
	\subfloat[MNIST dataset]{
	
	\begin{tikzpicture}[every mark/.append style={mark size=1pt}]
	
	\begin{axis}[width=8cm, height=5cm, legend columns=2, legend style={fill=none, font=\scriptsize, at={(0.5,1.03)},
		anchor=south}, xlabel= {$\log(S_\text{after})$}, ylabel= { $\log(L)$}]
	
	 
	\addplot [red, only marks]  table [x=logm, y=logloss]   {\tableone};
	\addplot [blue, only marks]  table [x=logmt, y=logloss]   {\tableone};
	
	\addplot [red, only marks, mark=x]  table [x=logm, y=logloss]   {\tabletwo};
	\addplot [blue, only marks, mark=x]  table [x=logmt, y=logloss]   {\tabletwo};
	
	\addplot [red, only marks, mark=diamond]  table [x=logm, y=logloss]   {\tablethree};
	\addplot [blue, only marks, mark=diamond]  table [x=logmt, y=logloss]   {\tablethree};
	
	\addplot [olive, solid, domain=0:22]{0.5*(ln(0.5*0.9*0.112/0.1))+0.5*x};

	\legend{$S$ on test 4-layer FC, $S$ on train 4-layer FC, $S$ on test 4-layer CNN, $S$ on train 4-layer CNN, $S$ on test VGG13, $S$ on train VGG13, Equation (\ref{eq:main})}

	\end{axis}

	\end{tikzpicture}}\qquad
	\subfloat[CIFAR-10 dataset]{
		
		\begin{tikzpicture}[every mark/.append style={mark size=1pt}]
		
		\begin{axis}[width=8cm, height=5cm, legend columns=2, legend style={fill=none, font=\scriptsize, at={(0.5,1.03)},
			anchor=south}, xlabel= {$\log(S_\text{after})$}, ylabel= { $\log(L)$}]
		
		
		\addplot [green, only marks]  table [x=logm, y=logloss]   {\tableonec}; 
		\addplot [teal, only marks, mark=triangle*]  table [x=logm, y=logloss]   {\tabletwoc}; 
		\addplot [blue, only marks, mark=triangle*]  table [x=logm, y=logloss]   {\tablethreec}; 
		\addplot [black, only marks]  table [x=logm, y=logloss]   {\tablefourc}; 
		\addplot [cyan, only marks]  table [x=logm, y=logloss]   {\tablefivec}; 
		\addplot [red, only marks, mark=o]  table [x=logm, y=logloss]   {\tablesixc}; 
		\addplot [pink, only marks, mark=o]  table [x=logm, y=logloss]   {\tablesevenc}; 
		
		\addplot [olive, solid, domain=-1:40]{0.5*(ln(0.5*0.9*0.2648/0.01))+0.5*x};

		\legend{4-layer CNN, VGG13, VGG16 XU, Alexnet, Alexnet XU, ResNet18 XU, ResNet18, Equation (\ref{eq:main})}

		\end{axis}

		\end{tikzpicture}}
		
		\caption{\textbf{(a)} Test loss versus sensitivity computed on the training set and the testing set for networks that are trained on 6000 samples of the MNIST training set. The Pearson correlation between $S$ computed on the training set and $S$ computed on the testing set is $\rho=0.9999$ and in the figure these two values meet each other at the exact same points. \textbf{(b)} Test loss versus sensitivity computed on the training set for networks that are trained on 1000 samples of the CIFAR-10 training set. We observe that the strong match between the empirical results and~(\ref{eq:main}) also holds for the sensitivity metric $S$ when it is computed on the training dataset. It is interesting to note that the y-axis is computed over the testing dataset, whereas the x-axis is computed without using the testing set, suggesting $S$ as a metric that does not require sacrificing the training samples for a validation set. In both figures the network parameters are initially drawn from the standard normal distribution unless otherwise stated as XU (Xavier technique with the uniform distribution).}
		\label{fig:test_vs_train}
		
	\end{figure*}

\end{document}